\documentclass[11pt]{article}

\usepackage[margin=1in]{geometry}
\usepackage{newtxtext}
\usepackage{microtype}
\usepackage{booktabs,tabularx,array}
\usepackage{amsmath,amssymb,amsthm}
\usepackage{graphicx}
\usepackage{float}
\usepackage{xcolor}
\usepackage{tikz}
\usetikzlibrary{arrows.meta,positioning}
\usepackage[section]{placeins}
\usepackage{hyperref}
\hypersetup{hidelinks}
\newtheorem{definition}{Definition}
\newcolumntype{Y}{>{\raggedright\arraybackslash}X}

\newcommand{\nll}{\mathcal{L}}
\newcommand{\stu}{\pi_{\theta}}
\newcommand{\tea}{\pi_{\psi}}
\definecolor{protocolblue}{HTML}{355C7D}
\definecolor{protocolorange}{HTML}{B66B31}
\definecolor{naivegray}{HTML}{979797}
\title{Streaming On-Policy Distillation:\\Decoupling Three Clocks for Stage-Wise Supervision Consumption}
\author{%
  Yun Zhu$^{1}$ \quad Da Wang$^{2}$\thanks{Corresponding author.}  \\[1ex]
  $^{1}$Independent Researcher, China \\
  $^{2}$School of Computer and Information Technology, Shanxi University, Taiyuan, China \\[0.5ex]
  \texttt{wanda@sxu.edu.cn}
}
\date{\today}

\begin{document}
\maketitle

\begin{abstract}
In on-policy distillation (OPD), response completion, supervision delivery,
and parameter updates need not share the same granularity. We develop
streaming OPD by decoupling these three clocks: a response repeatedly
delivers newly generated segments while generation continues, retaining
the full causal prefix at each delivery and computing loss only on new
positions. A consumption protocol then organizes supervision budgets,
position selection, staleness handling, and update scheduling. This
separates when generation supplies supervision from how training uses it,
allowing control over quantity, coverage, and version lag to change across
training stages. In GSM8K experiments distilling Qwen3-4B into Qwen3-0.6B
with VERL, supervised-budget matching reduces the early gap between best
test accuracies over seven common candidate checkpoints from 4.55 to 2.42
percentage points. Starting from the same asynchronous checkpoint,
synchronous continuation jointly resets the optimizer and switches consumption
schedules. Over seven common candidates at $u=384,416,\ldots,576$, its best
test accuracy is 73.39\%, versus 74.30\% for the synchronous baseline and
71.95\% for supervised-budget matching; the respective differences between
best scores are $-0.91$ and $-2.35$ percentage points. The unpaired extended
record reaches 76.12\% over eleven candidates, without a contemporaneous
synchronous reference. These observed test-set maxima are subject to
selection optimism; window means and full curves supplement them as process
statistics. Freshness filtering and disagreement--overlap selection improve
local peaks but do not establish sustained advantages or consistently close
the residual gap; tail-quota selection and vocabulary-support
interventions remain unpaired supplementary observations. A separate system
prototype shows lower data-supply waiting times, although its historical
runtime records do not establish
an equal-update speedup. These findings treat asynchrony as a selectable
consumption schedule rather than a fixed training mode: decoupling enables
continuous supply and independent batch assembly, while stage-wise
consumption uses that flexibility to balance quality and time. Quality
recovery and supply-side benefits are assessed separately; end-to-end gains
must also account for the continuation cost of reaching a target quality.
\end{abstract}

\section{Introduction}\label{sec:intro}

On-policy distillation trains a student with teacher feedback on responses
the student generates itself. Compared with learning from fixed teacher
responses, it supervises prefixes the student actually visits, helping
reduce the distribution mismatch between training and inference.
Long-horizon reasoning increases generation time and variation in response
duration, motivating systems that overlap generation, teacher scoring, and
parameter updates. Yet asynchronous execution alone does not answer a
separate question: \textbf{when should supervision within a response be
delivered, and at what granularity should it enter a gradient update?}

This question involves three clocks. Response completion follows generation
progress; supervision delivery follows the rule for releasing data units;
and parameter updates follow the trainer's batching and optimization rules.
Delivering complete responses couples the first two clocks. After switching
to segment delivery, batching by the number of delivered objects can still
make supervision per update depend on delivery granularity. Neither
asynchronous execution nor segmentation alone therefore establishes an
appropriate training organization. Our central idea is to separate these
three event sequences and explicitly control which supervision enters each
update.

Prior work already provides partial decoupling. AsyncOPD retains in-flight
prefixes while scoring completed samples and organizing asynchronous
updates~\cite{asyncopd}. TOPD generates and distills only the first $H$
positions; ADWIN updates on short windows and continues some prefixes in
the background to calibrate the window~\cite{topd,adwin}. A local OPD loss
does not depend on subsequent tokens, so a complete response is not a
prerequisite for computing it. We further change \textbf{the training role
of subsequently generated content within the same response}: rather than
training once on an early prefix or using continuation only as a probe, we
continually admit new segments as candidate supervision.

Streaming OPD implements this interface. Each delivery contains the prompt,
cumulative prefix, a new-segment mask, and version information. The full
prefix preserves the conditioning context, while only new positions
contribute to the current loss. Generation can continue after sealing a
segment, and the trainer assembles batches across segments instead of
equating one delivery with one update. A consumption protocol organizes
sealing, supervision budgets, position selection, staleness handling, and
synchronous or asynchronous scheduling. The contribution is not a new
distillation divergence, but an interface that makes generation supply and
gradient organization separately configurable.

Decoupling also creates a quality-control problem. Equal update counts may
consume different amounts of supervision; early-arriving segments and
later tails can alter position coverage; and generation and consumption
versions may differ. We describe \emph{consumption precision} along three
dimensions---quantity, coverage, and version lag---as the degree of control
the trainer exercises over supervision entering the gradient. It is not a
single score, nor does it assume that stricter selection is always better.
We use these consumption-side changes to diagnose quality costs without
attributing losses from different asynchronous settings to one established
cause.

The experiments extend this perspective from fixed configurations to
stage-wise training. Within seven common early candidate checkpoints,
supervised-budget matching reduces the gap between best test accuracies
from 4.55 to 2.42 percentage points, but a long-horizon residual remains.
Freshness filtering and disagreement--overlap selection improve local peaks
without establishing sustained advantages. In contrast, synchronous
continuation jointly resets the optimizer at the same asynchronous checkpoint
and switches consumption schedules. Across seven common candidates at
$u=384,416,\ldots,576$, its best test accuracy is 73.39\%, compared with
74.30\% for the synchronous baseline and 71.95\% for supervised-budget
matching, narrowing the best-score gap from 2.35 to 0.91 percentage points.
An unpaired extended record reaches 76.12\% over eleven candidates, without
a contemporaneous synchronous reference. These are observed test-set maxima,
not independently tested validation-selected models; means and full curves
remain process supplements. These results support a concrete path: use
asynchronous supply early, then adjust consumption to further improve
quality, rather than requiring one schedule throughout training.

System benefits and quality recovery must be considered together. Separate
measurements from a prototype show that finer-grained delivery can reduce
supply waiting under particular workloads. Its historical runtime comparison
has a discrepancy in completion counts, and the continuation trajectory
shows that quality improvement may require additional updates. Together,
these observations motivate the paper's design principle:
\textbf{decoupling creates a space of choices; stage-wise consumption
determines how to use it.} Asynchrony becomes a scheduling option, not an
end in itself that every training stage must maintain. Practical efficiency
depends on the full process of reaching target quality, not just the speed
of a local update.

\begin{samepage}
The contributions form three connected parts:
\begin{enumerate}
  \item \textbf{A continuously delivered training unit based on new
        segments within a response.} Full causal prefixes and new-position
        masks allow later content from the same response to repeatedly
        enter the candidate pool, distinguishing this interface from
        delivery after completion, one-time truncated training, and
        continuation probes.
  \item \textbf{A consumption-control framework that decouples completion,
        delivery, and updates.} Explicit sealing, budgets, selection,
        staleness handling, and scheduling distinguish context computation
        from actual supervision and allow consumption rules to change
        independently of individual deliveries and across training stages.
  \item \textbf{Empirical evidence for consumption diagnosis and
        stage-wise adjustment.} Budget experiments identify and repair part
        of the early gap; selection, freshness, quota, and local-objective
        experiments test alternative interventions. Synchronous continuation
        and extended observations show room for quality recovery from an
        asynchronous checkpoint, while a separate prototype records lower
        supply waiting times and pipeline runtimes.
\end{enumerate}
\end{samepage}

Section~\ref{sec:related} introduces the OPD loop and related approaches.
Section~\ref{sec:theory} develops the three-clock training interface.
Section~\ref{sec:exp} diagnoses supervision quantity, coverage, and version
lag; Section~\ref{sec:staged} examines the quality--efficiency tradeoff of
stage-wise continuation. Section~\ref{sec:limitations} discusses scope and
concludes.

\section{Background and Related Work}\label{sec:related}

\paragraph{Learning on the student's distribution.}
Classical knowledge distillation supervises the student with the teacher's
soft probability distributions~\cite{hinton}. Sequence-level distillation
instead approximates the sequence distribution with teacher beam-search
outputs and trains the student on those sequences~\cite{seqkd}. This differs
from obtaining per-position teacher feedback on prefixes visited by the
student itself. MiniLLM and GKD distill on student-generated data, applying
teacher feedback to prefixes visited by the student~\cite{minillm,gkd}. With complete
responses as the training unit, the basic loop is to generate responses
from prompts, obtain per-position teacher distributions on the same
prefixes, update the student using distributional differences, and pass
new parameters to generation. Synchronous execution waits for the current
round's data before updating; asynchronous execution overlaps generation
with training on delivered data. Complete-response delivery thus does not
imply synchronous execution of the entire system.

For example, given \textit{There are 3 bags with 4 apples each. How many
apples are there?}, a student might produce \texttt{There are 3 bags. Each
holds 4 apples. So there are 12 apples. <EOS>}. Complete-response delivery
waits until the response ends. Our interface delivers its three segments
successively, with the preceding context retained for the later two. We
use this example to explain delivery and updates; the segments are
illustrative, not actual tokenization boundaries or fixed sentence
boundaries. We retain teacher top-$k$ forward KL and study when this
supervision is delivered and how it enters updates, rather than replacing
the objective.

\paragraph{Asynchronous execution and data organization.}
IMPALA decouples experience generation from learning and uses V-trace to
address off-policy effects from lag between behavior and learner
policies~\cite{impala}. Policy mismatch thus requires treatment in its own
right in asynchronous systems. Its correction concerns reinforcement-learning
value estimation and actor--critic updates, rather than directly providing
a correction for forward KL on old prefixes here. AsyncOPD retains in-flight
prefixes across weight synchronization and continues generation, but its
teacher scoring consumes completed samples~\cite{asyncopd}. Near-Policy organizes asynchronous training through
response filtering and packing; $f$-OPD handles lag with freshness control,
anchor regularization, and buffer refresh~\cite{nearpolicy,fopd}. DivOPD
allocates budgets and selects data at the level of complete response turns
in multi-turn tasks~\cite{divopd}. These approaches already address
scheduling, staleness, and data selection, so we do not claim those
components as unique. We further make newly generated segments within a
single response into continuously deliverable training payloads.

\paragraph{Truncated prefixes and asynchronous continuation.}
TOPD generates and distills only the first $H$ tokens, while POPD gradually
expands the generation horizon, demonstrating that complete responses are
not required for OPD updates~\cite{topd}. Its segment ablations also select
different position intervals for loss computation after full generation.
ADWIN updates on the current prefix window while placing some unfinished
prefixes in an asynchronous probe pool; their continuations calibrate
future windows~\cite{adwin}. Early updates, partial-position masks, and
background continuation therefore all have precedents. Our distinction is
that \textbf{later new segments remain training data}: a response can be
sealed repeatedly, each segment enters the consumption pool with its
cumulative context, and the final segment follows the same delivery path.
Training is not confined to one prefix, and continuation is not used only
for auditing.

\begin{table}[H]
\centering\small
\caption{Mechanistic distinctions from neighboring methods. The comparison
concerns the organization of directly trained data and the use of later
content from the same response, not a performance ranking under matched
configurations.}
\label{tab:delivery}
\begin{tabularx}{\textwidth}{lYY}
\toprule
Method & Data used directly for training & Later content from the same response \\
\midrule
AsyncOPD~\cite{asyncopd} & Completed samples are teacher-scored and batched & In-flight prefixes are retained and continued; delivery follows completion \\
TOPD~\cite{topd} & The first $H$ positions are generated and distilled & The response is not continued beyond truncation for training \\
ADWIN~\cite{adwin} & Current-window prefixes support synchronous updates & Some prefixes continue in the background for window calibration \\
This work & New, prefix-preserving segments are repeatedly delivered and batched by budget & Later segments remain candidate supervision through the final segment \\
\bottomrule
\end{tabularx}
\end{table}

\paragraph{Position selection and reweighting.}
Prioritized experience replay adjusts sampling probabilities using
temporal-difference errors and applies importance weights to correct
nonuniform sampling bias~\cite{per}, illustrating that data availability
still leaves consumption frequency to be determined. Temporal-difference
error and teacher--student disagreement are different signals, however;
replay gains do not directly guarantee effective distillation-position
selection. RHO-1 selects positions using excess loss; TIP combines entropy
and teacher--student disagreement to characterize position importance;
TA-OPD organizes budgeted masks by teachability; and DFT reweights using
student probabilities~\cite{rho1,tip,taopd,dft}. Position selection and budget
constraints are thus not unique to this work. We treat them as consumption
variables on a segment stream and use random selection as a reference to
test whether reallocating supervision within a continually replenished
candidate pool can recover quality. Selection scores and the optimization
divergence are defined separately.

\paragraph{Prefix distributions and local compatibility.}
DAgger emphasizes the relationship between the distribution of training
states and the states a policy visits~\cite{dagger}. In autoregressive
generation, prefixes can be viewed as states, providing a perspective on
prefixes generated by older policies. We distinguish teacher--student
compatibility on a given prefix from the probability of visiting that
prefix, rather than interpreting local top-$k$ overlap as evidence that
stale data remain on-policy.

\paragraph{Curricula and stage-wise organization.}
Curriculum learning schedules examples by difficulty and learning progress;
self-paced learning selects examples according to the current model's loss
and gradually relaxes their inclusion criterion~\cite{curriculum,selfpaced}.
Both provide precedents for adjusting data organization over training.
Our control variables are the quantity, position coverage, version, and
scheduling of supervision entering updates, rather than a predefined
ordering of mathematics problems. Stage-wise consumption is related to
these ideas but is not equivalent to an easy-to-hard example curriculum.

\section{Method: Streaming OPD with Three Decoupled Clocks}\label{sec:theory}

Streaming OPD separates the end of generation, the availability of
supervision, and readiness to update. The \textbf{completion clock} marks
response termination or reaching the length limit. The \textbf{delivery
clock} marks the release of new segments under a sealing rule. The
\textbf{update clock} marks the trainer consuming supervision that meets
its batching conditions and updating parameters. These clocks are event
sequences, not three independent timers. Delivery requires the relevant
prefix to have been generated, and an update requires its supervision to
be ready. Decoupling removes the restrictions that only a complete response
can be delivered and that delivery granularity directly determines update
granularity.

We first explain why local supervision can be delivered early, then define
prefixes, masks, and versions, and finally organize the full loop with a
candidate pool and consumption protocol. These components turn the
mechanistic distinction in Table~\ref{tab:delivery} into an executable
training interface.

\subsection{Local Objectives Separate Completion from Delivery}

Given prompt $x$, the student generates a response $y=(y_1,\dots,y_T)$.
The forward-KL supervision from a fixed teacher at position $t$ is
\begin{equation}\label{eq:fkl}
  \nll_t(\theta)=D_{\mathrm{KL}}\!\left(
  \tea(\cdot\mid x,y_{<t})\,\|\,
  \stu(\cdot\mid x,y_{<t})\right).
\end{equation}
The conditioning prefix $y_{<t}$ has already been generated by the student;
later tokens do not enter this position's loss. Equation~\eqref{eq:fkl}
is the ideal full-vocabulary objective. The implementation instead sums
over a teacher-selected top-$k$ set $K_t^T$, without renormalizing teacher
or student probabilities within that set. Let $q_{t,z}$ and $p_{t,z}$ be
the teacher and student full-vocabulary probabilities on the same prefix,
and let $h_c(a)=\max(a,c)$. The implemented per-position objective is
\begin{equation}\label{eq:topk-impl}
  \widetilde{\nll}_t=
  \operatorname{clip}_{[0,C]}\!\left[
  \sum_{z\in K_t^T} e^{h_c(\log q_{t,z})}
  \bigl(h_c(\log q_{t,z})-h_c(\log p_{t,z})\bigr)\right].
\end{equation}
Here $c$ is a log-probability floor and $C$ a per-position loss ceiling.
The truncated sum can be negative; the implementation clips negative
values to zero. This approximation is therefore not equated with
full-vocabulary KL. Section~\ref{sec:setup} gives the configuration.
Truncation and clipping still require no future tokens: \textbf{complete
conditioning context is required for computation, but a complete response
is not required for delivery}. Streaming OPD exploits an existing property
of the local objective and changes when the system supplies it to the
trainer.

This differs from trajectory-level objectives that must wait for a terminal
reward. Process rewards or value bootstrapping can also give reinforcement
learning earlier feedback, but require the corresponding reward or
estimation mechanism. Here, early supervision comes directly from the
teacher distribution on a given prefix.

\subsection{Preserving the Prefix and Supervising Only New Segments}

Suppose a segment covers positions $a,\dots,b$. Its training input contains
the prompt and $y_{1:b}$, but its loss mask covers only the new segment:
\begin{equation}\label{eq:mask}
  m_t^{[a,b]}=\begin{cases}1,&a\leq t\leq b,\\0,&\text{otherwise}.
  \end{cases}
\end{equation}
The prefix $y_{<a}$ still participates in the forward pass but does not
contribute loss again in this sample; $y_{a:b}$ supplies the newly delivered
supervision. The final segment retains the termination marker actually
produced by generation. Sealing a segment must not be treated as ending a
response.

\begin{table}[!htbp]
\centering\small
\caption{Segment supervision within one response. The English fragments
illustrate the flow rather than actual tokenization boundaries; every row
retains the prompt. Each segment can enter the queue independently but
need not occupy a separate parameter update.}
\label{tab:example}
\begin{tabularx}{\textwidth}{lYYl}
\toprule
Delivery & Existing context (no loss) & New segment (supervised) & Segment mask \\
\midrule
Segment 1 & --- & \texttt{There are 3 bags.} & $1$ \\
Segment 2 & \texttt{There are 3 bags.} & \texttt{Each holds 4 apples.} & $0,1$ \\
Segment 3 & \texttt{There are 3 bags. Each holds 4 apples.} &
\texttt{So there are 12 apples. <EOS>} & $0,0,1$ \\
\bottomrule
\end{tabularx}
\end{table}

The distinction in Table~\ref{tab:example} is more than splitting a response
into three pieces. Complete-response delivery must wait for the final
segment containing \texttt{<EOS>}, whereas streaming delivery lets the first
two enter the pool earlier. When training on segment 2, segment 1 still
participates in attention but contributes no repeated loss. Segment 3
later arrives with both earlier segments as context. This preserves
conditioning information while separating the supervision assigned to
each delivery.

For fixed parameters, position sets, and weights, the sum of unreduced
per-position losses across segments equals the corresponding loss sum on
the complete response. If each segment is first reduced to a weighted
mean, those means must be combined using the total supervision weight in
each segment, not simply added. In actual training, parameters may change
between deliveries, as may the sampled position set, so local consistency
of the objective does not imply identical optimization trajectories.
Preserving prefixes makes the computation valid; the consumption protocol
below organizes the training process.

\subsection{Recording Versions: Local Compatibility versus Prefix Distributions}\label{sec:prefix}

Let $v_t$ be the parameter version that generated position $t$, and $v_u$
the version consuming it in update $u$. Define version lag as
\begin{equation}
  \mathrm{lag}_{u,t}=v_u-v_t.
\end{equation}
With a fixed teacher and recorded prefix, student updates do not change
the teacher target on that prefix, but do change the student's conditional
distribution and its probability of visiting the prefix. Version lag is
therefore a clue to distribution shift, not a direct measure of its size.

To describe local compatibility, we record the student's probability mass
on the teacher's top-$k$ set $K_t^T$ and the overlap between their top-$k$
sets:
\begin{equation}\label{eq:support}
  M_t=\sum_{z\in K_t^T}\stu(z\mid x,y_{<t}),\qquad
  O_t=|K_t^T\cap K_t^S|.
\end{equation}
Larger $M_t$ or $O_t$ indicates stronger local compatibility on the given
prefix. Neither measures the probability of visiting that prefix, so we
interpret these statistics together with training interventions when
assessing stale supervision.

\subsection{A Candidate Pool Connects Delivery and Updates}\label{sec:framework}

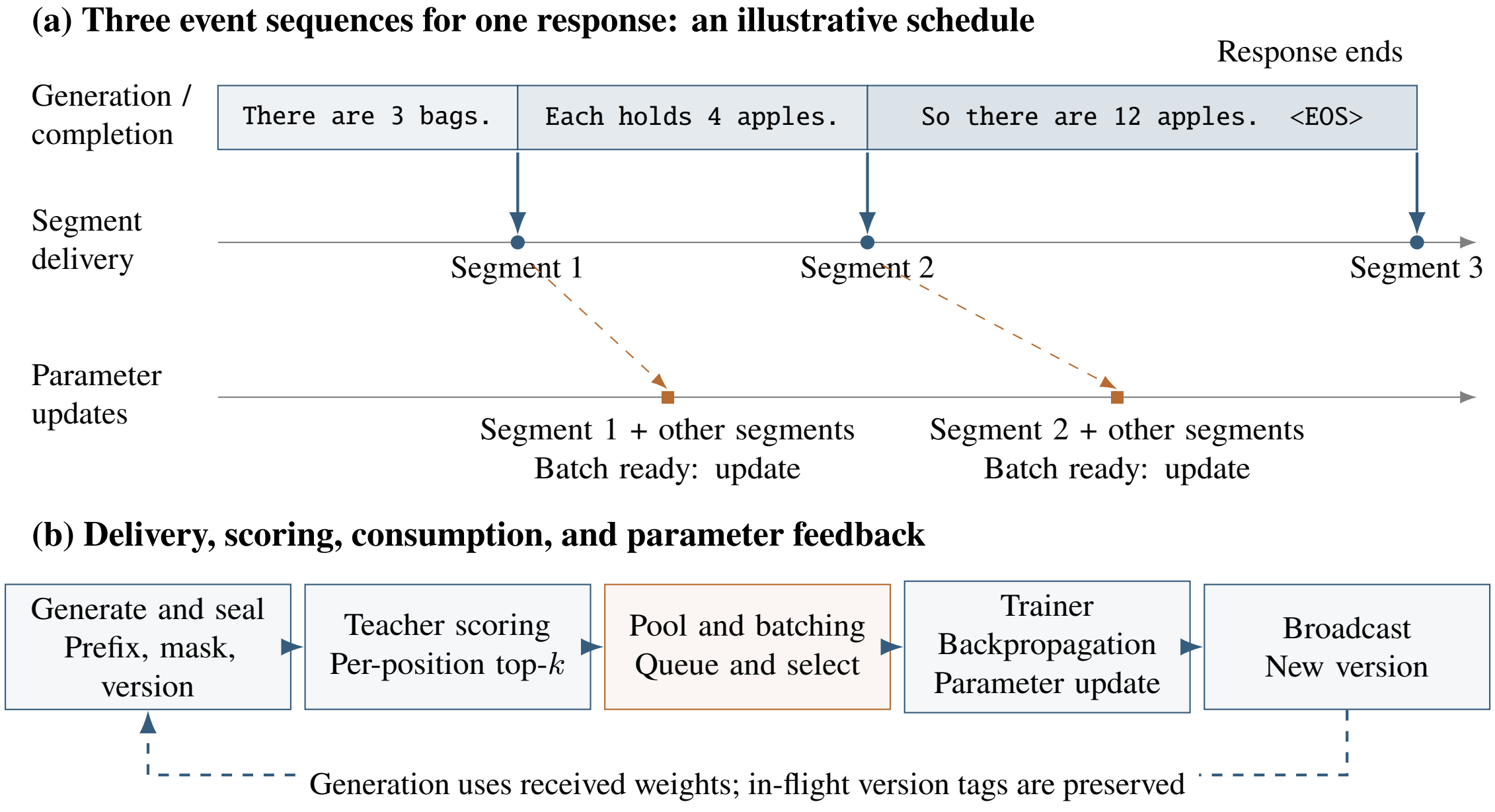
\begin{figure}[!htbp]
\centering
\begin{tikzpicture}[>=Latex,font=\small,
  stage/.style={draw=protocolblue,fill=protocolblue!5,
  text width=2.6cm,minimum height=1.25cm,align=center,font=\footnotesize},
  flow/.style={->,thick,draw=protocolblue}]
  \node[anchor=west,font=\small\bfseries] at (0,0.95)
    {(a) Three event sequences for one response: an illustrative schedule};
  \node[anchor=west,text width=1.85cm,font=\footnotesize] at (0,0) {Generation /\\completion};
  \draw[fill=protocolblue!8,draw=protocolblue] (2,-0.32) rectangle (5,0.32);
  \draw[fill=protocolblue!13,draw=protocolblue] (5,-0.32) rectangle (8.5,0.32);
  \draw[fill=protocolblue!18,draw=protocolblue] (8.5,-0.32) rectangle (14,0.32);
  \node[font=\scriptsize\ttfamily] at (3.5,0) {There are 3 bags.};
  \node[font=\scriptsize\ttfamily] at (6.75,0) {Each holds 4 apples.};
  \node[font=\scriptsize\ttfamily] at (11.25,0) {So there are 12 apples. <EOS>};
  \node[anchor=south east,font=\footnotesize] at (14,0.36) {Response ends};
  \node[anchor=west,text width=1.85cm,font=\footnotesize] at (0,-1.25) {Segment\\delivery};
  \draw[->,gray] (2,-1.25) -- (14.6,-1.25);
  \foreach \x/\n in {5/1,8.5/2,14/3} {
    \draw[flow] (\x,-0.35) -- (\x,-1.18);
    \fill[protocolblue] (\x,-1.25) circle (2pt);
    \node[below,font=\footnotesize] at (\x,-1.25) {Segment \n};
  }
  \node[anchor=west,text width=1.85cm,font=\footnotesize] at (0,-2.8) {Parameter updates};
  \draw[->,gray] (2,-2.8) -- (14.6,-2.8);
  \draw[->,dashed,protocolorange] (5.15,-1.48) -- (6.5,-2.72);
  \draw[->,dashed,protocolorange] (8.65,-1.48) -- (11,-2.72);
  \foreach \x in {6.5,11} {
    \fill[protocolorange] (\x-0.06,-2.86) rectangle (\x+0.06,-2.74);
  }
  \node[below,font=\footnotesize,align=center,text width=4.1cm] at (6.5,-2.87)
    {Segment 1 + other segments\\Batch ready: update};
  \node[below,font=\footnotesize,align=center,text width=4.1cm] at (11,-2.87)
    {Segment 2 + other segments\\Batch ready: update};
  \node[anchor=west,font=\small\bfseries] at (0,-4.2)
    {(b) Delivery, scoring, consumption, and parameter feedback};
  \node[stage] (roll) at (1.3,-5.3) {Generate and seal\\Prefix, mask, version};
  \node[stage] (teacher) at (4.3,-5.3) {Teacher scoring\\Per-position top-$k$};
  \node[stage,draw=protocolorange,fill=protocolorange!6] (pool) at (7.3,-5.3)
    {Pool and batching\\Queue and select};
  \node[stage] (train) at (10.3,-5.3) {Trainer\\Backpropagation\\Parameter update};
  \node[stage] (version) at (13.3,-5.3) {Broadcast\\New version};
  \draw[flow] (roll) -- (teacher);
  \draw[flow] (teacher) -- (pool);
  \draw[flow] (pool) -- (train);
  \draw[flow] (train) -- (version);
  \draw[flow,dashed] (version.south) -- ++(0,-0.65) -| (roll.south);
  \node[font=\footnotesize,fill=white] at (7.3,-6.7)
    {Generation uses received weights; in-flight version tags are preserved};
\end{tikzpicture}
\caption{Three clocks and the training loop. (a) An illustrative schedule,
not measured durations, fixed sealing boundaries, or a guarantee that every
segment enters an update before response completion. The final segment
also enters the pool for later batching. Dashed arrows indicate that a
segment may join supervision from other responses in an update; scoring
time is omitted. (b) The corresponding data and parameter flows. Full
prefixes and new-segment masks are detailed in Table~\ref{tab:example}.}
\label{fig:pipeline}
\end{figure}

A segment being deliverable does not mean it should immediately trigger an
update. The generator seals segments and supplies complete prefixes, masks,
and versions; the teacher scores the same prefixes; a candidate pool holds
supervision awaiting consumption. The trainer then assembles batches under
budget and selection rules, computes the loss, backpropagates, and updates
parameters. New versions are broadcast to generation, while in-flight data
retain their actual generation versions. Figure~\ref{fig:pipeline} connects
this loop to the three clocks: a response ends once but may deliver several
times, and an update may combine segments from different responses.

In the example from Table~\ref{tab:example}, segment 1 enters the candidate
pool after teacher scoring and is consumed only when it and other
candidates satisfy the batching conditions. Segment 2 arrives later with
segment 1 as context, but only the new positions corresponding to
\texttt{Each holds 4 apples.} count toward its supervision budget. Segment 3
follows the same path when generation ends. Pending supervision therefore
increases with delivery and decreases when batches are drawn; response
completion need not coincide with the delivery or update times of the first
two segments. The scoring service and queue can be deployed independently.

The current implementation seals at pause--resume boundaries caused by
parameter switching: it first delivers the nonempty segment generated
under the old version, then continues generation with the cumulative
prefix retained. When the response terminates or reaches its length limit,
it delivers the unreleased final segment. The basic scheme does not replay
previously delivered positions; completeness gating and replay are separate
interventions. Although this trigger is still tied to parameter broadcasts,
a delivered segment need not occupy its own update or wait for its response
to end. The early updates in the figure are scheduling opportunities
provided by the interface, not guarantees for every segment.

\subsection{Consumption Precision and Supervision Budgets}\label{sec:axes}

If batching continues to count delivered objects, replacing one complete
response with several segments can reduce supervision per update. Giving
priority to early-arriving segments can also change the position mix within
a window. A consumption protocol must therefore control what actually
enters the gradient, not merely accelerate generation. We characterize
consumption precision along three dimensions: \textbf{quantity}, the number
of loss-bearing positions per update and its variability;
\textbf{coverage}, the mix of positions across response lengths, segments,
and disagreement levels; and \textbf{version lag}, the interval between
generation and consumption versions. These dimensions can be recorded
separately but interact: masking old positions also changes quantity and
coverage. ``Precision'' denotes the granularity and degree of control,
not a new scalar metric or a claim that tighter control always improves
quality.

\begin{definition}[Consumption protocol]
A consumption protocol $\mathcal{C}=(\gamma,B,s,\rho,\sigma)$ consists of a
sealing rule $\gamma$, a per-update supervision budget $B$, a candidate
position-selection rule $s$, a staleness-handling rule $\rho$, and a
schedule $\sigma$ for synchronous or asynchronous data consumption.
The rule $\gamma$ may use segment length, version boundaries, or generation
termination; complete-response delivery is a special case.
\end{definition}

For candidate position $i\in P_u$ at update $u$, let $m_i\in\{0,1\}$ be its
original segment mask, $a_{u,i}\in\{0,1\}$ the selection result, and
$\rho_{u,i}\geq0$ an additional weight. With
$w_{u,i}=m_i a_{u,i}\rho_{u,i}$, the weighted objective is
\begin{equation}\label{eq:contract}
  \nll_u=\frac{\sum_{i\in P_u}w_{u,i}\nll_i}
  {\sum_{i\in P_u}w_{u,i}},\qquad
  N_u=\sum_{i\in P_u}\mathbf{1}[w_{u,i}>0].
\end{equation}
Actual computation replaces $\nll_i$ with the implemented objective in
Equation~\eqref{eq:topk-impl}. The budget constrains the \textbf{number of
positions participating in training}, $N_u$, while weights determine their
relative contributions. The supervised count equals the sum of weights
only with binary masking. Selection scores govern ranking or sampling and
are defined separately from loss weights. In segment-based samples,
inputs must also be distinguished from supervision: old prefixes consume
forward computation without supplying new loss positions. Raw-token
counts cannot therefore substitute directly for supervision budgets, and
repeated-prefix computation must be included in end-to-end time.

Batch assembly targets $B$, but segment boundaries can cause overshoot and
filtering can reduce final supervision. We therefore distinguish the target
budget, candidate count, and actual count. Our selection implementation
targets 160K candidate positions and retains 50\% within each microbatch,
corresponding to a nominal 80K supervision budget. Actual supervision is
measured after filtering; this is not a global ranking that selects exactly
the top 80K positions. Both synchronous and asynchronous schemes can use
the objective, with the schedule determining how the next batch is
obtained. Budgets and selection can thus be adjusted while retaining
segment supply, or consumption can switch to synchronous operation when
needed, without redefining the distillation loss.

\subsection{From Fixed Configurations to Stage-Wise Consumption}

A consumption protocol need not remain fixed throughout training. Writing
$\mathcal{C}_u$ for the protocol at update $u$ allows stage-wise changes to
budgets, selection, or data-consumption schedules while retaining the
teacher and local loss. A fixed protocol is a special case; a two-stage
schedule uses different configurations before and after a specified switch
point. This describes the interface's control capability, without assuming
that one schedule has higher precision or that all components must change
together.

\begin{table}[!htbp]
\centering\small
\caption{Principal consumption-protocol interventions. Each intervention
can affect several statistical dimensions and must be interpreted using
actual supervision counts and composition. Descriptive names associate
results with configurations; archival identifiers are mapped in
Appendix~\ref{app:counting}.}
\label{tab:design}
\begin{tabularx}{\textwidth}{lYYY}
\toprule
Variable & Direct change & Question tested & Intervention \\
\midrule
Budget $B$ & Supervision per update and its variability & Does matching supervision quantity narrow the gap? & Raw-token budget; supervised-budget matching \\
Selection $s$ & Positions retained within a budget & Are high disagreement or compatibility more useful? & High-disagreement selection; disagreement--overlap selection; uniform random selection \\
Staleness rule $\rho$ & Retained positions and relative weights & Does removing older-version supervision help? & Freshness filtering \\
Schedule $\sigma$ & Data consumption and its joint distribution & Can synchronous continuation narrow the residual? & Synchronous continuation with optimizer reset \\
\bottomrule
\end{tabularx}
\end{table}

We examine two levels of choice. First, we vary budgets, filtering, and
selection on an asynchronous segment stream to test which consumption
variables affect quality. Second, we continue from existing asynchronous
weights with synchronous consumption to assess whether a broader change
can further improve quality. Synchronous continuation implements the
latter two-stage path, jointly switching to synchronous consumption and
resetting the optimizer; its results therefore describe the combined
continuation procedure, not an isolated schedule change.
Table~\ref{tab:design} maps protocol variables to the experiments.

This formulation turns the three-clock separation into training decisions:
local losses determine which segments can be delivered, the consumption
protocol determines which supervision forms an update, and stage-wise
configuration determines when those rules change. We next diagnose quality
costs under fixed consumption modes, then examine changes to subsequent
consumption, without interpreting every intervention as evidence for the
same causal mechanism.
\section[Consumption-Side Diagnosis: Quantity Repair and Local Interventions]{Consumption-Side Diagnosis:\\Quantity Repair and Local Interventions}\label{sec:exp}

Finer-grained delivery changes the supervision entering each update. The resulting
quality gap therefore needs to be understood through the consumption process,
rather than solely through whether execution is asynchronous. This section
organizes the diagnosis around three questions: whether supervision is sufficient
in quantity, how to select positions that have already arrived, and whether the
delivery order needs to be adjusted. Changes to the update path and local
objectives provide additional comparisons. Supervised-budget matching reduces
the early best-score gap from 4.55 to 2.42 percentage points within common
candidate checkpoints, but the subsequent local interventions tested do not
consistently close the residual gap, despite some local peak improvements. This finding
motivates the staged choices in Section~\ref{sec:staged}: rather than only
optimizing a fixed rule, we adjust the consumption protocol in later training.

\subsection{Setup and Comparison Criteria}\label{sec:setup}

\textbf{Models and engines.} The student is Qwen3-0.6B and the teacher is
Qwen3-4B~\cite{qwen3}. We use a locally modified fully asynchronous trainer
based on VERL v0.8.0 and a vLLM~\cite{vllm} generation service;
HybridFlow~\cite{verl} is the underlying framework.
Training uses top-$k$ forward KL with a default learning rate of $10^{-5}$.
The generation temperature is 1.0 and top-$p$ is 1.0 during training, with
4 responses sampled per prompt and a maximum training response length of 2048.
Generation, teacher scoring, and training services are decoupled, and parameters
are broadcast at the synchronization interval.

\textbf{Objective implementation.} Equation~\eqref{eq:topk-impl} corresponds to
the FSDP2 path with direct backpropagation; the loss is not converted into PG
weights. The saved launch scripts specify $c=-10$ and $C=10$. The configuration
defaults to $k=128$ and mean reduction over supervised tokens; the main launch
scripts do not override these latter two settings. These details describe the
traceable implementation and launch recipes; they do not substitute for the final
configuration of each individual run. Vocabulary-level objective interventions
are discussed separately in Section~\ref{sec:diagnostics}.

\textbf{Main experiments.} The synchronous baseline is compared with naive
streaming, which uses segment-wise consumption. The raw-token budget
collects an 80K budget measured in raw tokens, whereas supervised-budget matching
collects a budget measured in supervised tokens. The reduced learning rate
retains the sample-count-based batching of naive streaming and only reduces
the learning rate to $0.58$ times its default value, without enabling a token
budget. Synchronous continuation switches from the checkpoint obtained with
supervised-budget matching at $u=352$ to synchronous consumption and resets the
optimizer. Configurations and evaluation results for position-selection
experiments are reported separately in Section~\ref{sec:select}; all archival
method and branch identifiers are mapped in Appendix~\ref{app:counting}.

\textbf{Training data and resources.} The main experimental launch configurations
use a training file mixing GSM8K~\cite{gsm8k} and DAPO~\cite{dapo} mathematics
prompts, with source fields \texttt{openai/gsm8k} and \texttt{math\_dapo},
respectively. These citations identify data resources, not adoption of the
DAPO optimization algorithm; the source fields alone do not establish use
of the full original resources or identical filtering. In trainer/generator/
teacher order, the launch configurations allocate 4/1/1 GPUs to the synchronous
baseline and naive streaming, and 2/1/1 GPUs to the raw-token budget, reduced
learning rate, and supervised-budget matching. These numbers describe the launch
recipes, not a uniform resource configuration across all continuation branches.
Quality comparisons align candidate trainer-update checkpoints, not necessarily
the selected updates; we do not use these allocations to claim an efficiency
advantage under equal GPU budgets.

\textbf{Training axis.} Let $u$ denote the trainer update count and $g$ the index
of the broadcast parameter version. In this set of experiments, parameters are
broadcast every four updates. The main runs satisfy $u=4g$. Synchronous continuation
and the late-start position-selection experiments start from the weights obtained
with supervised-budget matching at $u=352$ and reset the optimizer. Their
cumulative continuation update count is $r=4g$, and their total update count is
$u=352+r$. The correspondence between directory and log counts, including
separately identified branches that must not be connected across restarts, is
given in Appendix~\ref{app:counting}.

\textbf{Evaluation.} The GSM8K test set contains 1319 questions. Evaluation uses
temperature 0 and a common procedure for numerical answer extraction and equality
checking. The main experimental evaluation entry point sets the maximum generation
length to 1024, distinct from the training limit of 2048. Length settings for the
historical multi-benchmark diagnostics are distinguished separately. Cross-machine
reevaluation statistics are provided in Appendix~\ref{app:counting}.

\textbf{Common-candidate reporting.} Each experimental group uses a fixed set
of common trainer-update checkpoints. The primary quality result is the
\emph{best observed test accuracy}: each method independently takes its maximum
within this candidate set, with ties assigned to the earliest update $u^*$.
Differences compare these independently selected best scores with the
synchronous baseline's best score; they are not same-update paired differences.
This aligns the candidate range, evaluation opportunities, and update ceiling,
not necessarily the selected updates or actual GPU costs. The fixed sets are
$u=32,64,\ldots,224$ for the seven-point early comparison (excluding $u=256$);
$u=256,288,\ldots,576$ plus $596$ for the twelve-point long-horizon comparison;
$u=384,416,448$ for late-start position selection; and
$u=384,416,\ldots,576$ for the seven-point synchronous-continuation comparison.

Selection is performed within each run first; where multiple runs are available,
their selected scores are averaged with equal weight, never by picking the best
run. The independent delivery-timing screen has only one candidate, $u=16$,
per run, so its reported run averages are unchanged. Its unequal run counts and
common synchronous anchor do not constitute paired seeds. Each of the other
main comparisons has one trajectory per method per group; branch records do
not create additional main entries. Accuracy (\%) and differences (percentage
points) are reported to two decimal places, with differences computed before
rounding. Window means are retained as process summaries, with mean differences
also computed before rounding; subtracting displayed means may differ by 0.01.
Selected updates and window means appear in Appendix~\ref{app:statistics}.

Maximizing test accuracy introduces selection optimism: these results are not
validation-set model selection followed by independent testing. Neither best
scores nor means establish statistical equivalence. Means and full curves
describe the correlated evaluation process, not independent seeds or sustained
performance at the peak. Results are not ranked across configuration groups.
Records without reliable update mappings and extended observations without a
contemporaneous synchronous reference remain unpaired supplements, excluded
from common-candidate baseline differences.

\textbf{Auditing.} We record per-update supervision counts and their coefficient
of variation (standard deviation divided by the mean), KL quantiles, the student's
probability mass on the teacher's top-$k$ set, and set overlap counts, grouped by
version lag and segment type. We refer to segments carrying the
\texttt{is\_terminal} flag as ``terminal segments.'' The fraction of supervised
positions in these segments is not the fraction accounted for by the single EOS
token. Historical short-horizon screening, objective-implementation diagnostics,
and the system prototype use different configurations, as described in their
respective subsections; they are not combined with the main experiments into a
unified single-factor ablation.

\subsection{Diagnostic Path and Evidence Overview}\label{sec:evidence-map}

We organize the diagnosis by different dimensions of consumption, juxtaposing
observed differences with corresponding interventions.
Table~\ref{tab:evidence-map} places the observations, tested interventions, and
outcomes side by side. At the first level, we adjust budgets, position selection,
and delivery policies while retaining asynchronous consumption, testing which
repairs are available within a fixed consumption mode. Comparisons involving the
update path and vocabulary-level objectives distinguish changes to data supply
from changes to the objective. At the second level, we retain existing
asynchronous weights and adjust the consumption mode in a subsequent phase,
examining quality recovery and its cost. This organization places local repairs
within fixed rules and protocol changes between training phases in a single chain
of evidence.

\begin{table}[!htbp]
\centering\small
\caption{Evidence overview for Sections~\ref{sec:exp} and~\ref{sec:staged}.
Each row uses its own configuration and candidate window; best scores are
selected independently within common candidates where mappings are available.
Results are neither ranked across groups nor added together as causal
contributions.}
\label{tab:evidence-map}
\begin{tabularx}{\textwidth}{YYYY}
\toprule
Dimension & Observations & Tested intervention & Main finding \\
\midrule
Supervision quantity & Segmenting reduces per-update supervision and increases its variability & Raw-token budget; supervised-budget matching & Early best-score gap decreases from 4.55 to 2.42 percentage points \\
Update scale & Both supervision quantity and cross-update variability change & Reduced learning rate & Early best test accuracy 71.49\%, 0.07 percentage points above supervised-budget matching \\
Position coverage & Larger gaps on medium-length and long responses; a lower share of supervision in terminal segments & Tail-quota selection for positions that have arrived & Unpaired five-point best 68.01\%; no baseline difference without a step map \\
Version lag & Stale positions retain disagreement, with similar local probability coverage & Freshness filtering & Best 72.78\% versus 71.95\% for supervised-budget matching, but wins only the first of three points \\
Position selection & Many candidate positions have little local disagreement & High-disagreement selection; disagreement--overlap selection; uniform random selection; teacher-support mass reweighting & Random best 71.80\% versus high-disagreement 69.45\%; disagreement--overlap best 72.55\% \\
Delivery timing & Tail-segment availability and consumption order may affect updates & Completeness gating; terminal-first scheduling; stale-segment replay; regeneration with current weights & At the sole candidate per run, mean improvements for gating and regeneration, but neither reaches the synchronous anchor \\
Objective implementation & The same supervised positions can correspond to different update objectives & PG weighting; dynamic vocabulary truncation; bidirectional rank divergence & Specific PG implementations degenerate; vocabulary changes remain unpaired observations \\
\midrule
Staged consumption & A quality gap remains after quantity repair & Synchronous continuation with an optimizer reset & Best 73.39\% versus 74.30\% synchronous and 71.95\% supervised-budget matching; best-score differences $-0.91$ and $-2.35$ percentage points \\
\bottomrule
\end{tabularx}
\end{table}

Naive streaming's best test score trails the synchronous baseline's best by
4.55 percentage points within the seven early common candidates. We next
examine which consumption-side adjustments can narrow this gap.

\subsection{Supervision Quantity: Restoring the Per-Update Training Signal}\label{sec:tax-exists}\label{sec:qty}

After segmentation, the first quantity to examine is the number of positions that
actually contribute to the loss in each update, rather than the number of delivered
objects. The audit summary in Table~\ref{tab:audit} reports approximately
46--48K supervised tokens per update for naive streaming, compared with
approximately 74.8K for the synchronous baseline; the former also has greater
cross-update count variability. Repeated prefixes increase input length but add no
new supervised positions. Even when the loss is normalized by the token count,
sampling coverage and gradient estimates still change with the actual amount of
supervision.

\begin{figure}[!htbp]
\centering
\includegraphics[width=\textwidth]{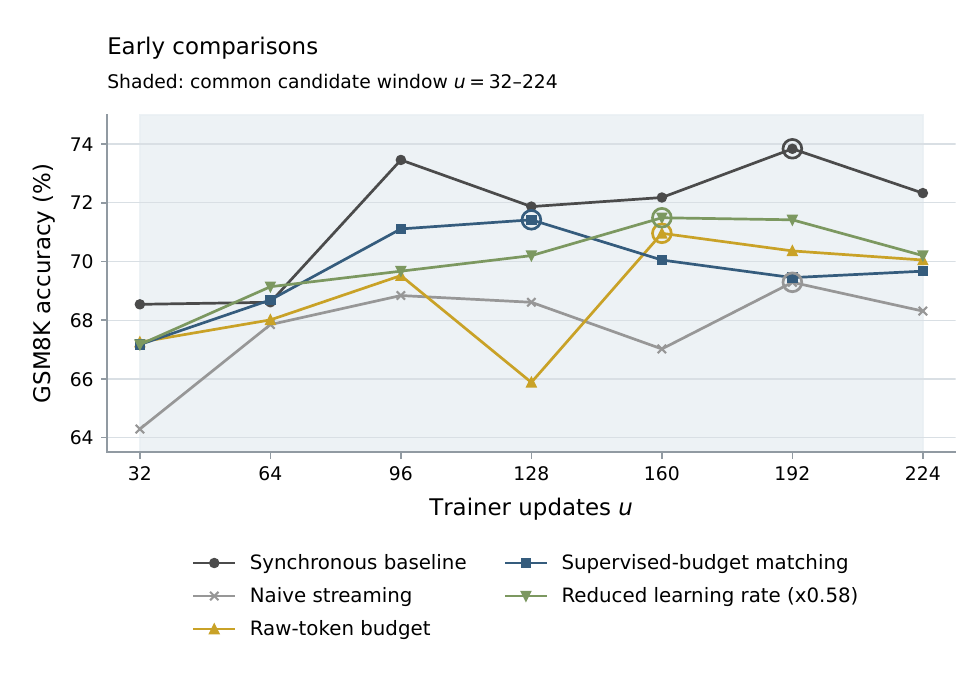}
\caption{Early budget and learning-rate comparisons. Shading covers the seven
common candidates $u=32,64,\ldots,224$, corresponding to
Table~\ref{tab:equal-update}; open circles highlight the earliest maximum for
each method within this set. Solid lines connect measured values without
smoothing. Supervised-budget matching and the reduced learning
rate narrow naive streaming's best-score gap, but their subsequent trajectories differ.}
\label{fig:early}
\end{figure}

\begin{table}[!htbp]
\centering\small
\caption{Consumption-side audit, retaining source precision. Supervised-token
counts are approximate range/value summaries; coefficients of variation are
reported range/value summaries; terminal-segment shares are approximate
values. Intervals are source-reported ranges, not extrema over a common
window or confidence intervals. These are count and composition audits,
not accuracy means. Definitions are given in Appendix~\ref{app:statistics}.}
\label{tab:audit}
\begin{tabularx}{\textwidth}{Ycc}
\toprule
Metric (summary value) & Naive streaming & Synchronous baseline \\
\midrule
Supervised tokens per update & Approx.\ 46--48K & Approx.\ 74.8K \\
Coefficient of variation of supervision counts across updates & 0.43--0.65 & 0.045 \\
Share of supervised positions in terminal segments & Approx.\ 40\% & Approx.\ 80\% \\
\bottomrule
\end{tabularx}
\end{table}

On this basis, we construct budgets using raw-token counts and supervised-position
counts, respectively, retaining naive streaming as a reference.
Table~\ref{tab:equal-update} shows a best-score gap of 4.55 percentage points
between naive streaming and the synchronous baseline over the seven common
candidates. Introducing a budget narrows this gap, and supervised-budget
matching has a higher peak than the raw-token budget (Figure~\ref{fig:early}).
This indicates that local losses
available before completion need to be designed jointly with batching rules suited
to their supervision granularity.

\begin{table}[!htbp]
\centering\small
\caption{Early comparison within common candidates
$u=32,64,96,128,160,192,224$. Each method independently selects its best
test accuracy from one trajectory; differences compare best scores, not
same-update pairs. Selected updates and process means are in
Table~\ref{tab:checkpoint-audit-early-long}. Reduced learning rate retains
naive streaming's sample-count batching and uses $0.58$ times the default
learning rate, without a token budget.}
\label{tab:equal-update}
\begin{tabularx}{\textwidth}{Y>{\raggedleft\arraybackslash}p{2.7cm}>{\raggedleft\arraybackslash}p{3.6cm}}
\toprule
Method & Best test accuracy (\%) & Difference between best scores vs. baseline (pp) \\
\midrule
Synchronous baseline & 73.84 & $0.00$ \\
Naive streaming & 69.29 & $-4.55$ \\
Raw-token budget & 70.96 & $-2.88$ \\
Supervised-budget matching & 71.42 & $-2.42$ \\
Reduced learning rate & 71.49 & $-2.35$ \\
\bottomrule
\end{tabularx}
\end{table}

\paragraph{Budget repair and the long-horizon residual gap.}
Within the same early candidate set, supervised-budget matching reaches
71.42\%, versus 69.29\% for naive streaming and 73.84\% for the synchronous
baseline, reducing the best-score gap from 4.55 to 2.42 percentage points.
The raw-token budget reaches 70.96\%, with a best-score difference of
$-2.88$ percentage points, possibly because it counts context prefixes that
contribute no new loss; however, this comparison does not separately isolate
the causal contribution of that factor. The results support tying the budget
to \emph{actual supervised positions}, rather than approximating supervision
solely by input length. As a process supplement, the window means are 69.65\%
for supervised-budget matching, 67.74\% for naive streaming, 71.55\% for the
baseline, and 68.86\% for the raw-token budget; their mean differences are
reported in Table~\ref{tab:checkpoint-audit-early-long}.

Reducing the learning rate provides another intervention path. The reduced
learning rate retains naive streaming batching without enabling a token budget
and uses $0.58$ times the default learning rate. Its early best accuracy is
71.49\%, with a baseline best-score difference of $-2.35$ percentage points,
0.07 percentage points above supervised-budget matching's peak. Its process
mean is 69.90\%. The two interventions yield similar early gains but do not
have identical subsequent trajectories. Supplementary pointwise records,
including the reduced learning rate's low intermediate score and later
recovery, are retained in Appendix~\ref{app:counting}; they do not alter the
fixed candidate set. Early results or a single checkpoint are insufficient
to establish the mechanism through which insufficient supervision acts.

\begin{figure}[!htbp]
\centering
\includegraphics[width=\textwidth]{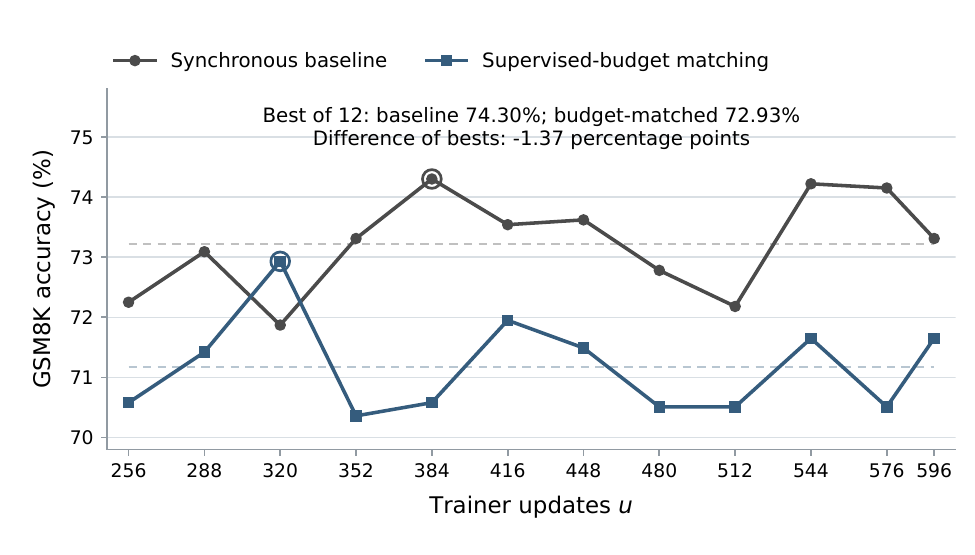}
\caption{Long-horizon curves over twelve common candidates
$u=256,288,\ldots,576$ plus $596$. Open circles highlight each method's
earliest maximum; top annotations report best scores and differences between
them. Solid lines connect measured values without smoothing. Horizontal
dashed lines show process means, not confidence intervals. Supervised-budget
matching reaches 72.93\% versus the baseline's 74.30\%, a best-score gap of
1.37 percentage points despite exceeding it at individual checkpoints.}
\label{fig:long-paired}
\end{figure}

These results support treating the supervised-token budget as a basic constraint
for segment-wise training, rather than continuing to approximate the amount of
training by the number of delivered objects. Figure~\ref{fig:long-paired} shows
twelve long-horizon common candidates: the synchronous baseline reaches
74.30\%, naive streaming 71.57\%, and supervised-budget matching 72.93\%,
with best-score differences of $-2.73$ and $-1.37$ percentage points for the
latter two. The corresponding process means are 73.22\%, 69.85\%, and
71.18\%; supervised-budget matching's mean gap is 2.04 percentage points.
Quantity repair alone is thus insufficient to attain synchronous quality.
The following analysis turns to the composition and versions of the positions
entering the budget, rather than merely increasing the nominal number of
training updates. Long-horizon and early candidate windows are reported
separately; neither their best scores nor their means provide a cross-window
causal decomposition.

\subsection{Within-Pool Adjustments: Coverage, Versions, and Position Selection}\label{sec:comp}\label{sec:select}

After quantity repair, the next question is which positions to retain within the
budget. Keeping asynchronous data supply and the original distillation objective,
this subsection compares position composition, version age, and local disagreement
with corresponding interventions: tail quotas test coverage adjustments, freshness
filtering tests the removal of old positions, and ranking and reweighting test
local scores. All operate on candidate positions that have already arrived and
cannot directly add subsequent segments that have not yet been generated.

Table~\ref{tab:cells} summarizes the rules and common-window outcomes.
High-disagreement selection, disagreement--overlap selection, and uniform
random selection target a 160K candidate budget and retain 50\% of positions
within each microbatch. Freshness filtering uses a nominal 80K budget,
teacher-support mass reweighting only reweights positions, and tail-quota
selection adds a tail quota to uniform random selection. Actual supervision
quantity is counted according to the rules in Section~\ref{sec:axes}.
Tail-quota selection lacks a reliable update map and is reported separately
as an unpaired observation below, not included in the matched-window table.
We first present the observations motivating each intervention and then
interpret its results.

\begin{table}[!htbp]
\centering\small
\caption{Late-start GSM8K comparison within the three common candidates
$u=384,416,448$. Each method independently selects its best test accuracy
from one trajectory, not three seeds. Differences compare best scores with
the synchronous baseline, not same-update pairs. Selected updates and process
means are in Table~\ref{tab:checkpoint-audit-late-cont}; full trajectories
and branch records remain in Appendix~\ref{app:selection}.}
\label{tab:cells}
\begin{tabularx}{\textwidth}{>{\raggedright\arraybackslash}p{3.5cm}Y>{\raggedleft\arraybackslash}p{1.8cm}>{\raggedleft\arraybackslash}p{2.8cm}}
\toprule
Method & Intervention rule & Best test accuracy (\%) & Difference between best scores vs. baseline (pp) \\
\midrule
Synchronous baseline & Synchronous consumption & 74.30 & $0.00$ \\
Supervised-budget matching & Budget measured in supervised positions & 71.95 & $-2.35$ \\
High-disagreement selection & Retain the top 50\% by per-position FKL & 69.45 & $-4.85$ \\
Disagreement--overlap selection & Rank by FKL $\times$ top-$k$ overlap & 72.55 & $-1.75$ \\
Uniform random selection & Retain 50\% uniformly at random & 71.80 & $-2.50$ \\
Freshness filtering & Mask positions with $\mathrm{lag}>0$ & 72.78 & $-1.52$ \\
Teacher-support mass reweighting & Reweight by the student's probability mass on the teacher's top-$k$ set, without removing positions & 70.74 & $-3.56$ \\
\bottomrule
\end{tabularx}
\end{table}

\subsubsection{Position Coverage: From Length Groups to Tail Quotas}

\paragraph{Larger gaps on medium-length and long responses.}
At each checkpoint, we divide responses into quartiles according to the generation
lengths of the synchronous baseline, with Q1 through Q4 ordered from shortest to
longest. At $u=448$, supervised-budget matching trails the baseline by 7.70
percentage points in Q3. At $u=480$, the differences for Q2, Q3, and Q4 are
$-2.70$, $-3.90$, and $-4.90$ percentage points, respectively. These grouped
diagnostics retain the source summaries' one-decimal precision despite the
two-decimal display; they are not substitutes for overall best scores within
the common candidate set. The two checkpoints suggest that medium-length and long responses are
affected regions deserving closer inspection. Length grouping alone cannot
attribute errors to content, formatting, or stopping mechanisms.

Generation length provides another comparison. At $u=448$, the baseline, naive
streaming, and supervised-budget matching have mean generation lengths of 266,
283, and 253 tokens, respectively; at $u=596$, the corresponding values are 270,
283, and 260 tokens, all reported as integers. Supervised-budget matching generates
shorter responses than the baseline, yet an accuracy gap remains, indicating that
length inflation alone is insufficient to explain the gap. The audit summary also shows a decline in the share of
supervision in terminal segments, from approximately 80\% to approximately 40\%.
This metric, however, depends on the granularity of segment labeling and is not
the EOS-token share. Supervision coverage on medium-length and long trajectories
deserves attention, but whether deferring tail segments causes quality loss still
needs to be distinguished from a mere change in statistical accounting.

\paragraph{Tail-quota selection: an unpaired supplementary observation.}
To test position composition directly, tail-quota selection targets a 40\% share
of tail positions with $t\ge160$ in the selected budget, sampling uniformly within
each stratum and otherwise using the same configuration as uniform random
selection. This target comes from the reference share of approximately 0.40 in
synchronous supervision; the corresponding share in the asynchronous candidate
pool is approximately 0.22. The best test accuracy among the five reported
checkpoints is 68.01\%, with a process mean of 66.06\%. Their update map is
not reliable, so this unpaired observation is excluded from
Table~\ref{tab:cells} and from common-candidate baseline differences;
it is not ranked against a different window of uniform random selection.
The low pointwise scores and original range remain in Appendix~\ref{app:selection}.

A quota can only redistribute positions that have already arrived; it cannot add
tail segments that have not yet arrived. Increasing the tail share may also change
the version and difficulty distributions of the selected positions. Matching the
tail share therefore does not imply matching the overall consumption distribution.
The current observations do not establish a quality recovery and support
distinguishing \emph{supply coverage} from \emph{within-pool redistribution},
but low scores alone cannot establish that older tail positions are the cause
or rule out other quota schemes.

\subsubsection{Version Lag: From Fresh--Stale Grouping to Freshness Filtering}\label{sec:stale}

Tail quotas may also change the version composition of the selected positions.
We therefore need to examine separately whether fresh and stale positions have
different local statistics and whether removing stale positions improves quality.

\paragraph{Stale positions retain disagreement, with similar local probability coverage.}
Table~\ref{tab:posaudit} distinguishes quantile summaries for the full early
candidate pool of uniform random selection from grouped summaries for fresh and
stale positions. The
per-update KL medians over the first 8 updates are reported as approximately
0.0015--0.002. This indicates that many positions in these updates have little
disagreement, not that they have strictly zero gradients or no training value.
In the fresh--stale grouping, stale positions have slightly higher mean KL but
similar top-$k$ probability mass. Supervision from older versions therefore cannot
simply be treated as ineffective data.

\begin{table}[!htbp]
\centering\small
\caption{Audit of uniform random selection's early candidate pool, retaining
source precision. The first row is a reported range of per-update KL medians,
and the second an approximate summary of per-update mass 10th percentiles,
over the first 8 updates. The last two report grouped mean KL and approximate
mass summaries for fresh and stale positions at approximately update 8.
These are local-distribution audits, not accuracy means. Reduction conventions
are given in Appendix~\ref{app:statistics}.}
\label{tab:posaudit}
\begin{tabularx}{\textwidth}{YYY}
\toprule
Metric & Summary value & Supported interpretation \\
\midrule
Per-update median KL & Approx.\ 0.0015--0.002 & Many positions have little disagreement \\
Per-update 10th percentile of mass & Approx.\ 0.997 & High local probability coverage \\
Mean KL of stale/fresh positions & 0.14 / 0.12 & Residual disagreement remains in the stale group \\
Mass of stale/fresh positions & Approx.\ 0.997 / 0.997 & Similar local probability coverage \\
\bottomrule
\end{tabularx}
\end{table}

\paragraph{Freshness filtering yields no consistent advantage.}
Freshness filtering masks all positions with $\mathrm{lag}>0$ under a nominal 80K
supervised-token budget. Over the three common candidates $u=384,416,448$,
its best test accuracy is 72.78\%, compared with 71.95\% for supervised-budget
matching, a peak improvement of 0.83 percentage points. Its best-score gap
to the synchronous baseline is 1.52 percentage points (Table~\ref{tab:cells}).
This local improvement is real in the recorded curve but not sustained:
freshness filtering exceeds supervised-budget matching only at the first
of the three points. The process means are 71.49\% and 71.34\%, respectively.
The earlier six-point summary and separate branch records remain in
Appendix~\ref{app:selection} as supplementary observations, not additional
main comparisons.

The aligned-window results indicate that removing stale positions is insufficient
to resolve the residual gap. Its effect still needs to be understood in conjunction
with the post-filtering supervision quantity, position composition, and optimizer
state.

Two sets of historical high-lag records yield opposite rankings for freshness
filtering after uniform rescoring, and complete run mappings are unavailable.
They are reported as supplementary observations in
Appendix~\ref{app:diagnostics}. Version age should therefore be considered jointly
with actual supervision quantity and position composition, rather than used as a
standalone criterion of value.

\subsubsection{Position Selection: Local Disagreement Does Not Imply Supervision Benefit}

\begin{figure}[!htbp]
\centering
\includegraphics[width=\textwidth]{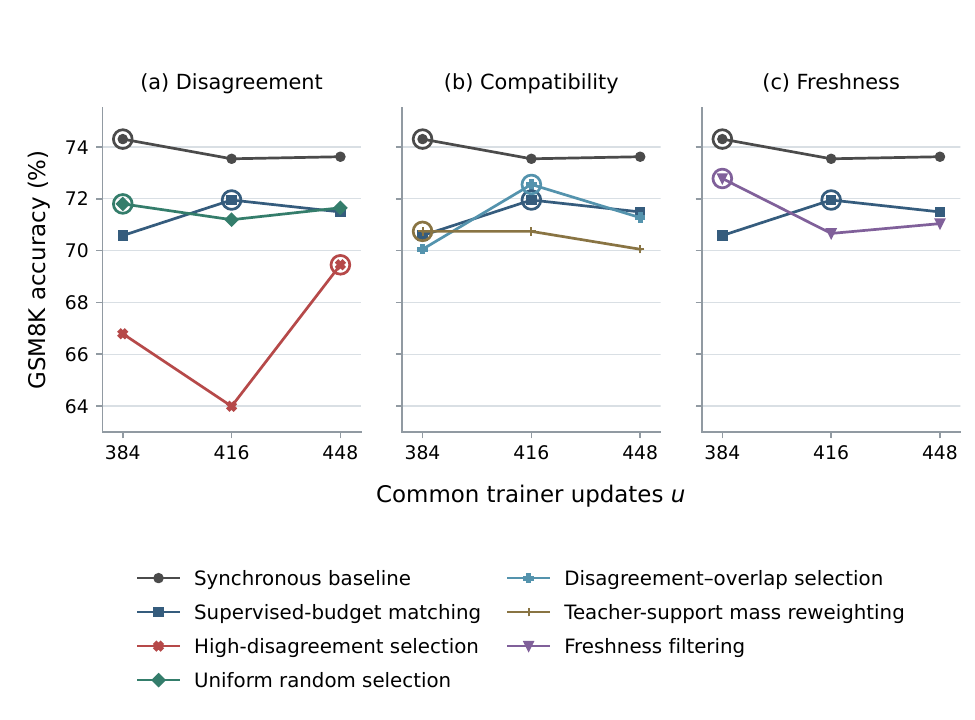}
\caption{Late-start GSM8K comparisons at the three common checkpoints
$u=384,416,448$: (a) high-disagreement selection and uniform random selection;
(b) disagreement--overlap selection and teacher-support mass reweighting;
(c) freshness filtering. All panels use the same x- and y-axis ranges and
repeat the same synchronous-baseline and supervised-budget-matching curves.
Lines connect measured points only, without interpolated checkpoints or
smoothing; open circles highlight each method's earliest maximum within
these candidates. Table~\ref{tab:cells} reports these best scores and their
differences; full mapped trajectories appear only as supplementary evidence
in Figure~\ref{fig:curves-full}.}
\label{fig:curves}
\end{figure}

The candidate-pool audit also shows that many positions have small KL values.
This motivates prioritizing high-disagreement positions, but disagreement only
measures the difference between teacher and student on the current prefix; it
does not directly specify the learning benefit of that position. We therefore
compare high-disagreement selection by pure KL ranking, disagreement--overlap
selection by compatibility-assisted ranking, and teacher-support mass reweighting,
using uniform random selection under the same nominal budget as a reference.

\paragraph{High-disagreement selection performs poorly.}
High-disagreement selection reaches a best test accuracy of 69.45\% over
$u=384,416,448$, a best-score difference of $-4.85$ percentage points from
the synchronous baseline. Its MATH500 endpoint, whose parent dataset is
MATH~\cite{math}, has an accuracy of 29.00\% ($n=500$), with a mean generation
length of 566 and a truncation rate of 24.8\%, suggesting that this branch
also has substantial length-control problems.
This single-checkpoint result on a different benchmark is not included in
the GSM8K comparison.

\paragraph{Uniform random selection outperforms high-disagreement selection.}
Over the same three candidates, uniform random selection reaches 71.80\%,
compared with 69.45\% for high-disagreement selection, and is higher at each
of the three points. Its peak remains 0.15 percentage points below
supervised-budget matching's 71.95\%. Disagreement--overlap selection reaches
72.55\%, improving on the budget-matched peak by 0.60 percentage points,
whereas teacher-support mass reweighting reaches 70.74\%. Thus, the two
compatibility-based interventions do not have the same outcome: overlap
selection improves the local peak, but the curve does not establish a
sustained advantage. As process supplements, the means are 71.55\% for
random selection, 66.74\% for high-disagreement selection, 71.29\% for
disagreement--overlap selection, and 70.51\% for mass reweighting.
Figure~\ref{fig:curves} compares
exactly this shared window. Full pointwise records, including low scores
outside the window, remain in Appendix~\ref{app:selection}.

The three types of within-pool adjustment provide complementary evidence:
the unpaired tail-quota observation does not establish quality recovery,
removing stale positions yields no consistent advantage, and high KL does
not automatically correspond to high benefit. They show that local statistics
cannot replace a holistic examination of the actual consumption distribution.
We next move the intervention from within the candidate pool to the delivery
process, examining the effects of changing segment availability and consumption
order.

\subsection{Delivery Timing: Gating, Replay, and Regeneration}\label{sec:timing}

Within-pool rules can only process positions that have already arrived, whereas
delivery policies can change which segments the trainer receives and when.
Short-horizon screening compares completeness gating, terminal-first scheduling,
stale-segment replay, and regeneration with current weights to examine whether
these adjustments can narrow the gap to a synchronous anchor. This set uses a
separate configuration and a short observation window. Stale-segment replay
and regeneration with current weights also change content and computation,
so they cannot be treated as timing-only ablations.

\begin{table}[!htbp]
\centering\small
\caption{Independent short-horizon scheduling screen. Values are run means
at $u=16$; differences use the synchronous anchor of 66.79\%. Run-level
records and aggregation rules appear in Appendix~\ref{app:diagnostics}.}
\label{tab:timing}
\begin{tabularx}{\textwidth}{Y>{\raggedleft\arraybackslash}p{2.2cm}>{\raggedleft\arraybackslash}p{3.0cm}}
\toprule
Method & Mean accuracy (\%) & Difference vs. synchronous baseline (pp) \\
\midrule
Synchronous baseline & 66.79 & $0.00$ \\
Naive streaming & 59.06 & $-7.73$ \\
Completeness gating & 61.79 & $-5.00$ \\
Terminal-first scheduling & 53.15 & $-13.64$ \\
Stale-segment replay & 55.76 & $-11.03$ \\
Regeneration with current weights & 62.70 & $-4.09$ \\
\bottomrule
\end{tabularx}
\end{table}

\paragraph{Scheduling adjustments offer limited improvement.}
At $u=16$, the synchronous baseline averages 66.79\%, compared with 59.06\%
for naive streaming. Completeness gating averages 61.79\% and regeneration
with current weights 62.70\%, both above naive streaming; terminal-first
scheduling and stale-segment replay average 53.15\% and 55.76\%, respectively.
None reaches the synchronous anchor (Table~\ref{tab:timing}). The mean
baseline differences range from $-4.09$ percentage points for regeneration
with current weights to $-13.64$ for terminal-first scheduling. Thus, there
is room for improvement in the delivery process, but these tested policies
have not eliminated the gap within this window.

\subsection{Objective-Side Comparisons: Update Paths and Vocabulary Support}\label{sec:diagnostics}

The preceding interventions primarily change how supervision enters an update.
Another class of modifications changes how optimization proceeds once supervision
has entered the update. To distinguish these two levels, this subsection examines
implementations that use FKL values for PG weighting, as well as local objectives
that modify vocabulary support and the form of the divergence. These are not
single-factor variants of the data-supply protocol. Their results test whether
objective changes provide another route to quality improvement.

\paragraph{The update path has independent failure modes.}\label{sec:estimator}
Both implementations that use FKL values for PG weighting produce 0.00\% accuracy
and degenerate outputs. With clipping, the clipping fraction rises to approximately
97\%; without clipping, the gradient norm rises from approximately 29 to 81.
Rescoring leaves the results unchanged. This shows that updates based on FKL
values passed through importance ratios and clipping differ substantively from
direct FKL optimization. This set uses a reduced mini-batch size and diagnoses
the mechanisms of specific implementations; it is not combined with the main
experiments to compute effect sizes.

\paragraph{Loss-shape interventions on vocabulary support.}
Position-selection experiments redistribute the budget along the \emph{sequence
position} dimension. Another class of interventions changes the support set and
divergence form at each supervised position along the \emph{vocabulary} dimension.
The two vocabulary-level variants reuse the initial asynchronous checkpoint of
uniform random selection, target a 160K candidate budget, and retain 50\% of
positions uniformly within each microbatch. Dynamic vocabulary truncation
uses knee-point detection to determine a dynamic $k$ at each position within the
teacher's top-$128$ support, retaining only the forward-KL terms for the
corresponding vocabulary items. Following Switch-KD~\cite{switchkd}, bidirectional
rank divergence uses the teacher and student to select dynamic support
sets separately. In each of its two branches, it aligns distributions obtained by
applying temperature-scaled softmax to logit differences, and sums the losses of
the two reverse-KL branches.

These are unpaired supplementary observations: dynamic vocabulary truncation
has a four-checkpoint best test accuracy of 68.61\% and a process mean of
67.57\%; bidirectional rank divergence has a three-checkpoint best of 64.22\%
and a process mean of 63.49\%. Reliable update mappings are unavailable,
so neither enters a common-candidate baseline difference or a ranking against
uniform random selection. Original ranges and all pointwise values remain
in Appendix~\ref{app:selection}.

For dynamic vocabulary truncation, one possible explanation is that truncating
the support reduces supervisory information about relations among candidate
vocabulary items: the
same number of supervised positions no longer corresponds to the same scope of
teacher feedback. Once the support changes, loss values also depend on the number
of summands and the normalization convention. A smaller loss does not automatically
imply closer agreement over the full vocabulary distribution. Whether the lost
information is predominantly in the tail should be assessed using the distribution
of dynamic $k$ and evaluations on a common support over the same batch of prefixes,
not merely by comparing the magnitudes of the modified losses.

By contrast, bidirectional rank divergence changes support selection, the objects
being aligned, and the divergence form simultaneously. One possible effect is that within-set relative
relations may not adequately constrain the allocation of probability between the
set and vocabulary items outside it. For example, for a fixed selected set, adding
the same constant to all logits in that set leaves within-set differences
unchanged but can change the set's total probability mass. This concerns a change
in the subset relative to its complement, not a uniform shift of the entire
vocabulary, which already leaves softmax probabilities unchanged. Coverage and
overlap of the two support sets affect this constraint. Moreover, reverse KL here
acts on distributions of logit differences, so interpretations based on
vocabulary probabilities cannot be applied directly. This analysis clarifies that
the two objectives impose different constraints, but it does not yet allow the
observed low scores to be attributed solely to changes in probability mass,
support selection, or KL direction.

Unlike this isolated transfer of one component, Switch-KD targets multimodal
distillation and also includes autoregressive cross-entropy, conventional
alignment, and visual switch distillation. These results therefore cannot be
extrapolated into an assessment of its complete method. The role of additional
supervision, such as cross-entropy, also requires separate testing.

Neither of the two vocabulary-level modifications tested here has shown a quality
benefit. Together with the PG diagnostics, which use a different configuration,
they underscore that changing the update path or a local objective cannot be
regarded as merely changing data supply. The quality improvements observed in
this paper include supervised-token budgeting, local peak gains from
freshness filtering and disagreement--overlap selection, and subsequent
consumption adjustments. Vocabulary-level interventions remain a separate
diagnostic direction.

\paragraph{Review of scoring conventions.}
Collapse judgments caused by answer-extraction errors in the early screening have
been corrected through rescoring, whereas the degenerate outputs in the PG
diagnostics persist. These two phenomena must be distinguished. The corresponding
scores, coverage, and correction process are documented in
Appendix~\ref{app:diagnostics}.

\subsection{From Local Interventions to Staged Choices}\label{sec:local-to-staged}

This section yields two connected insights. First, supervision quantity after
segmentation must be controlled explicitly: supervised-budget matching reduces
the early best-score gap from 4.55 to 2.42 percentage points and provides a
basis for continued training. Second, freshness filtering and
disagreement--overlap selection improve local peaks without establishing
sustained advantages, while high-disagreement selection performs poorly.
These interventions do not consistently close the residual gap; the unpaired
tail-quota and vocabulary observations do not establish a corresponding
quality recovery. Short-horizon delivery policies offer only
partial improvement, and the PG diagnostics expose separate failure modes.
Supervision-consumption precision therefore cannot simply be equated with
filtering strength or any single local statistic.

This motivates the next level of design: rather than requiring one fixed
consumption rule throughout training, we examine whether existing asynchronous
weights can benefit from protocol adjustments in later phases. The failure of
the tested local interventions to consistently close the residual gap does not
imply that the corresponding dimensions have no effect. Staged choices further
test whether extending consumption control from individual updates to the training
process can improve quality while continuing to use existing weights.
\section[Stage-Wise Consumption: Quality Recovery and Efficiency Trade-offs]{Stage-Wise Consumption:\\Quality Recovery and Efficiency Trade-offs}\label{sec:staged}

Section~\ref{sec:exp} examined local remedies within a fixed asynchronous consumption regime. This section extends control across training stages: starting from existing weights, we adjust the protocol governing subsequent consumption. We examine three questions in turn: whether the quality gap can be narrowed within a matched-update window, whether extended training brings further improvement, and how additional training affects the time savings. Synchronous continuation jointly introduces synchronous consumption and an optimizer reset; the analysis below evaluates this complete continuation path rather than attributing its effects to a single scheduling factor.

\subsection{Adjusting Subsequent Consumption: Narrowing the Matched-Window Gap}\label{sec:a3}

Synchronous continuation starts from the checkpoint obtained with supervised-budget matching at $u=352$, resets the optimizer, and switches to synchronous consumption, while supervised-budget matching retains asynchronous training. The two paths share the same weights at the switch and subsequently perform their respective gradient updates. The question is not whether retraining from scratch succeeds, but whether a model obtained through an asynchronous phase can approach the quality of fully synchronous training by adjusting subsequent consumption. The synchronous baseline is aligned by total update count.

\paragraph{Configuration comparison before and after the switch.}
The launch recipe for supervised-budget matching seals segments at version switches, collects an 80K budget counted over actual supervised positions, retains both fresh and stale positions with a stale-position weight of 1, and applies no additional filtering by disagreement score. The synchronous continuation records confirm initialization from the merged model weights at $u=352$, optimizer reinitialization rather than restoration of the previous optimizer state, and synchronous consumption. However, the final settings for segment sealing, the budget, and filtering are not fully listed. The term ``synchronous'' alone therefore cannot establish whether data were still drawn as segments or had instead been batched as complete responses. Accordingly, synchronous continuation is evaluated as a joint continuation path, not as a single-factor intervention that changes only $\sigma$ while holding all other protocol components fixed.

\begin{figure}[!htbp]
\centering
\includegraphics[width=\textwidth]{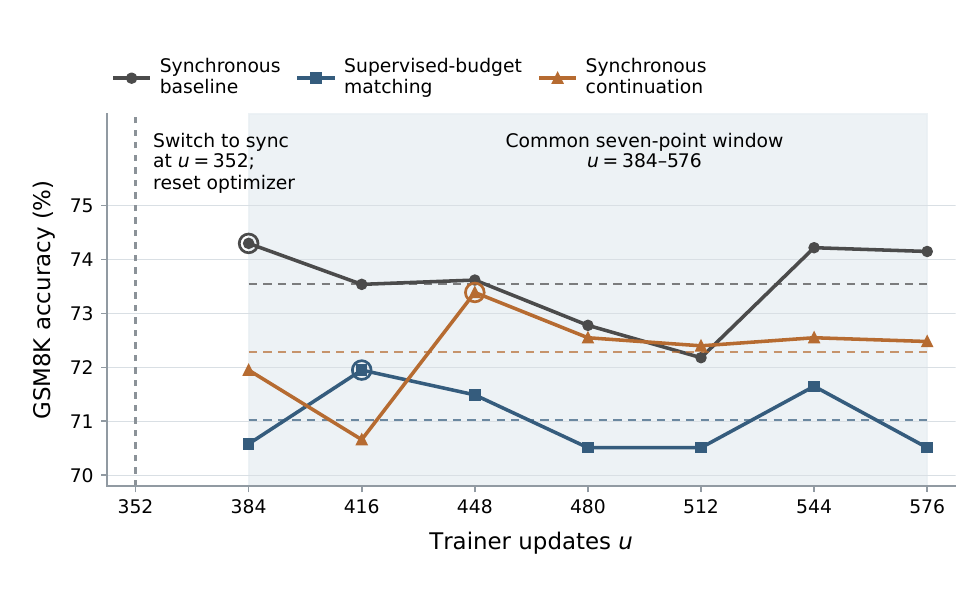}
\caption{Synchronous-continuation curves over seven common candidates
from $u=384$ to $576$ at 32-update intervals. The vertical line marks the switch and
optimizer reset at $u=352$; shading includes the first two post-switch
evaluations. Open circles highlight each method's earliest maximum.
Synchronous continuation reaches 73.39\%, supervised-budget matching 71.95\%,
and the synchronous baseline 74.30\%; the respective best-score differences
from the baseline are $-0.91$ and $-2.35$ percentage points. Lines connect
measured values without smoothing; horizontal dashed lines show seven-point
process means, not confidence intervals.}
\label{fig:a3-paired}
\end{figure}

\begin{figure}[!htbp]
\centering
\begin{tikzpicture}[x=1cm,y=0.64cm,font=\small]
  \node[align=center,text width=6.6cm,font=\small\bfseries] at (3.35,1.1) {(a) Early budget intervention\\$u=32$--$224$};
  \node[align=center,text width=6.6cm,font=\small\bfseries] at (10.75,1.1) {(b) Synchronous\\continuation comparison\\$u=384$--$576$};
  \foreach \k in {0,1,2,3,4,5} {
    \draw[gray!25] (0.6,-\k) -- (6.1,-\k);
    \draw[gray!25] (8.0,-\k) -- (13.5,-\k);
    \node[left,font=\footnotesize] at (0.6,-\k) {\ifnum\k=0 $0$\else $-\k$\fi};
    \node[left,font=\footnotesize] at (8.0,-\k) {\ifnum\k=0 $0$\else $-\k$\fi};
  }
  \draw[thick] (0.6,0) -- (6.1,0);
  \draw[thick] (8.0,0) -- (13.5,0);
  \fill[naivegray] (1.4,0) rectangle (2.8,-4.55);
  \fill[protocolblue!80] (4.0,0) rectangle (5.4,-2.42);
  \fill[protocolblue!80] (8.8,0) rectangle (10.2,-2.35);
  \fill[protocolorange!80] (11.4,0) rectangle (12.8,-0.91);
  \node[below] at (2.1,-4.55) {$-4.55$};
  \node[below] at (4.7,-2.42) {$-2.42$};
  \node[below] at (9.5,-2.35) {$-2.35$};
  \node[below] at (12.1,-0.91) {$-0.91$};
  \node[anchor=north,align=center,text width=3.0cm] at (2.1,-6.0) {Naive\\streaming};
  \node[anchor=north,align=center,text width=3.0cm] at (4.7,-6.0) {Supervised-budget\\matching};
  \node[anchor=north,align=center,text width=3.0cm] at (9.5,-6.0) {Supervised-budget\\matching};
  \node[anchor=north,align=center,text width=3.0cm] at (12.1,-6.0) {Synchronous\\continuation\\(optimizer reset)};
  \node[rotate=90,align=center,text width=3.3cm,font=\footnotesize] at (-1.0,-2.5) {Best-score gap from the\\synchronous baseline\\(percentage points)};
\end{tikzpicture}
\caption{Two common-candidate comparisons of consumption-side interventions.
(a) Supervised-budget matching reduces the early best-score gap from 4.55 to
2.42 percentage points over seven common candidates. (b) Synchronous
continuation with an optimizer reset narrows the best-score gap from 2.35
to 0.91 percentage points over $u=384,416,\ldots,576$. Each method selects
its maximum independently; these are not same-update paired differences.
The panels use different candidate windows and do not provide a cross-window
decomposition of causal contributions.}
\label{fig:interventions}
\end{figure}
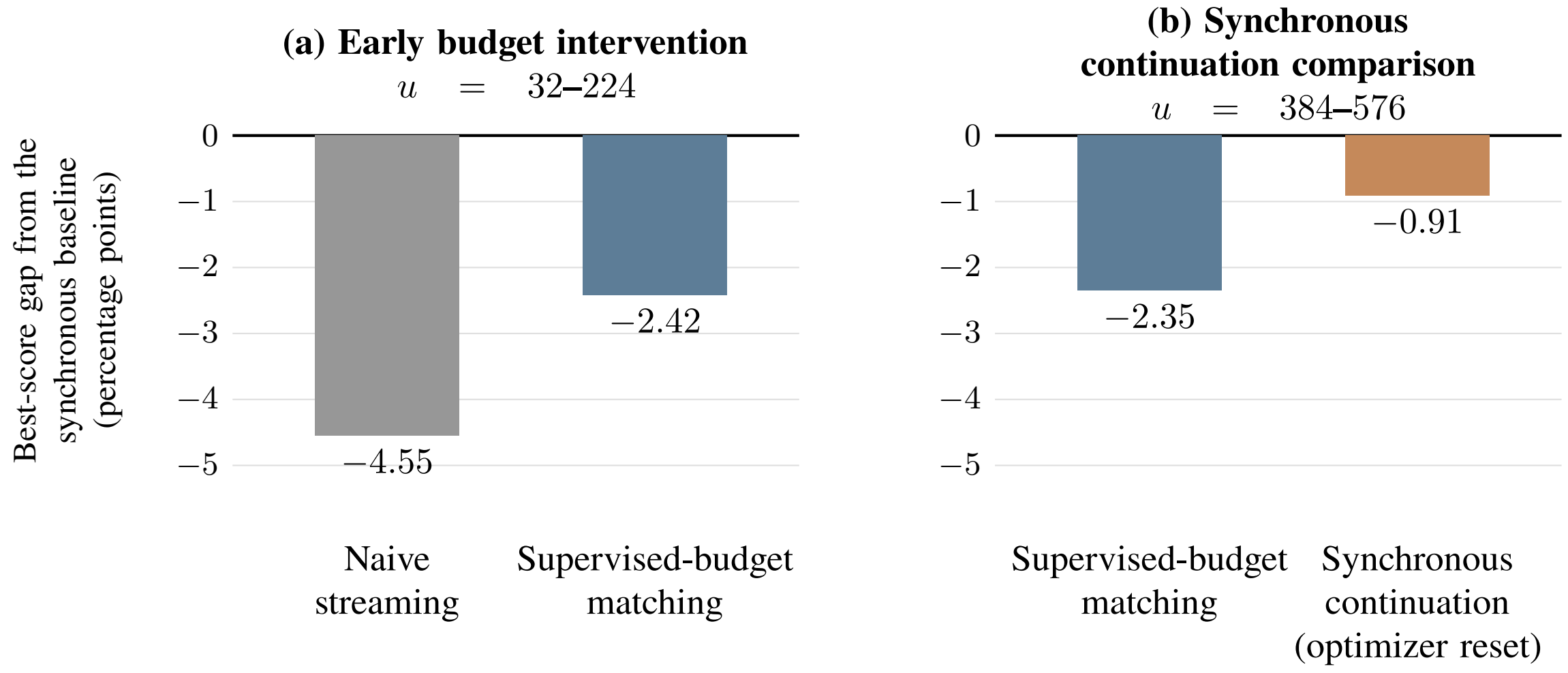

Figure~\ref{fig:a3-paired} covers all seven common candidates after the switch
($u=384,416,\ldots,576$), retaining the first two rather than excluding them
because of their lower scores. Best test accuracies for the synchronous
baseline, supervised-budget matching, and synchronous continuation are
74.30\%, 71.95\%, and 73.39\%, respectively. Synchronous continuation narrows
the best-score gap from 2.35 to 0.91 percentage points within this common
candidate set and update ceiling; the selected updates need not coincide.
As process supplements, the respective means are 73.54\%, 71.03\%, and
72.28\%, with mean differences of $-2.51$ and $-1.26$ percentage points for
the latter two. Figure~\ref{fig:interventions} juxtaposes the best-score
comparison with the early budget intervention: the latter adjusts the amount
of supervision under the same data-supply regime, whereas synchronous
continuation starts from existing weights and jointly adjusts consumption
and optimizer state.

The matched window supports a narrowing of the gap, not a claim that
synchronous continuation has consistently caught up. Weights obtained through
an asynchronous phase can nevertheless approach synchronous quality through
subsequent training, motivating further examination of this two-stage path.
The historical five-point summary and pointwise residuals remain in
Appendix~\ref{app:continuation} and do not replace the seven-point main
comparison. Quality changes over longer training require an extended
observation window; synchronous records outside the matched window are also
listed separately in that appendix.

\subsection{Extended Observation: Further Improvement after a Short Plateau}\label{sec:extended}

The matched-update window shows quality recovery. The next question is whether
this improvement continues over longer training rather than consisting solely
of a short-lived fluctuation after the switch. The extended synchronous
continuation record restarts at cumulative stage update $r=224$ (total update
$u=576$) and contains eleven checkpoints at $r=224,256,\ldots,544$, or
$u=576,608,\ldots,896$. Its best observed test accuracy is 76.12\% at
$r=480$ ($u=832$), with an eleven-point process mean of 74.06\%. This is an
\emph{unpaired extended observation}: it has no contemporaneous synchronous
reference and is not pooled with the seven-point main comparison. In
particular, the records at $u=576$ contain distinct values of 72.48\% and
72.40\%, which are retained separately rather than spliced or averaged.
Appendix~\ref{app:continuation}, including Figure~\ref{fig:a3-extended}, preserves
the complete trajectory, restart mapping, and historical subwindow statistics.

\paragraph{The short plateau and the observation window.}
The extended trajectory shows a short plateau at $u=576$--$672$ before
later improvement; this is distinct from the first two post-switch evaluations
at $u=384,416$. This \emph{apparent latency} indicates that continuation
benefits may appear only later in the evaluation curve; the absence of sustained improvement within
a short window does not imply that further training has no room for recovery.
Sparse checkpoints cannot identify the precise onset of improvement or a
generally applicable waiting period. The pointwise chronology and its latency
interpretation are retained in Appendix~\ref{app:continuation}. Evaluation of a
continuation strategy must consider both the switching point and the
observation window.

\paragraph{Quality recovery and equal-budget comparison.}
The extended trajectory contains multiple higher observations, rather than a
single isolated peak, but they require additional training. The matched
seven-point window supports a narrowing of the baseline gap, whereas the
unpaired extended record supports further improvement after a plateau.
Without a contemporaneous synchronous reference, the latter does not establish
that synchronous continuation catches up under an equal update budget or
demonstrate statistical equivalence. Historical synchronous scores from other
runs and windows remain supplementary records in Appendix~\ref{app:continuation},
not substitutes for a matched reference.

\paragraph{A possible explanation involving joint adaptation of supply and policy.}
Switching the consumption regime changes the amount of supervision, the composition of positions, and the version distribution entering the gradient; subsequent parameter updates also change the prefixes visited by the student. Continuation may therefore involve both adaptation to the new consumption distribution and further capability learning. If a mismatch between supply and policy affects the rate of recovery, then, all else being equal, a larger preswitch distribution shift should be associated with a longer recovery process, while consumption control during the transition might shorten it. This is a testable prediction from a ``drift--alignment'' perspective, not a mechanism established by the post-plateau improvement. We examine only one initial switching point and cannot yet distinguish the contributions of new learning, correction of existing biases, and evaluation fluctuations. Nor can these observations establish that recovery cost increases monotonically with staleness.

\subsection{Efficiency Evidence: Data-Supply Benefits in a Separate Prototype}\label{sec:gains}

The preceding two subsections provide quality evidence for a narrowing of the matched-window gap and further improvement with extended training, respectively. The benefits of extension, however, also require additional training. Evaluating the entire path requires examining how much waiting time finer-grained delivery can save during the initial phase and whether these savings can offset subsequent costs. This subsection first reports data-supply and runtime measurements from a separate prototype; Section~\ref{sec:cost} then discusses overall accounting for the two-stage strategy.

The systems value of segmented delivery lies in creating additional opportunities to supply data from unfinished responses and allowing the consumption protocol to organize those opportunities. An asynchronous complete-response pipeline can already overlap generation and training. If its queue remains nonempty, earlier delivery need not further reduce training time. Our additional refinement concerns the within-response unit of data supply, enabling segment sealing, budgets, position selection, and update scheduling to be configured separately over a stream of segments. The controllability of this interface is therefore the design contribution; waiting time and end-to-end runtime under a particular workload test whether that flexibility translates into practical benefits.

The prototype summary reports a supply wait for a batch of 128 samples of 47--218 seconds with complete-response delivery, compared with approximately 0.2 seconds with sealed-segment delivery. Another pair of runs targeting 64 updates reports runtimes of approximately 55 and 35 minutes, respectively (Table~\ref{tab:system}), but their completion fields read 63/64 and 64/64. It is not yet possible to determine whether this discrepancy reflects a numbering convention or a difference in the number of updates actually completed. The reported runtime ratio of approximately $1.6$ is therefore not treated as a verified equal-update speedup. The sealed-segment run also discarded segments because of queue overflow, with a reported drop rate of approximately 2.9\%. Appendix~\ref{app:statistics} gives the denominator and the results after subsequent queue expansion.

Data-supply latency and runtime describe local waiting and pipeline execution, respectively; a proportional reduction in the former cannot be directly converted into end-to-end acceleration. Waiting-time measurements under the tested workload support the systems value of finer-grained delivery, while historical runtimes provide leads for further cost accounting. They complement the preceding evidence on quality recovery but are not combined into a claim of quality-preserving acceleration for the stage-wise strategy.

\begin{table}[!htbp]
\centering\small
\caption{Separate prototype measurements, retaining source precision. Each
row uses its stated statistical basis rather than a common accuracy window:
range/approximation, p95, peak summary, count/upper bound, value/range, or
approximate runtime. $\approx$ indicates an approximation, intervals are
reported ranges, and $\leq$ denotes a reported upper bound. Target update
counts do not resolve the recorded completion-count discrepancy; these
measurements do not establish equal-update speedup. Details appear in
Appendix~\ref{app:statistics}.}
\label{tab:system}
\begin{tabularx}{\textwidth}{YYYY}
\toprule
Metric & Reporting basis & Complete-response delivery & Sealed-segment delivery \\
\midrule
Supply wait for a 128-sample batch & Range/approximation & 47--218 s & $\approx0.2$ s \\
Data-supply wait & p95, target: 128 updates & 5.06 s & 0.34 s \\
Trainer idle fraction & Peak summary & 57\% & $\approx10\%$ \\
Lag-1 tokens/batch & Count/reported upper bound & 18,073 & $\leq743$ \\
Fresh-token fraction & Value/reported range & 0.74 & 0.87--0.95 \\
Runtime & Target: 64 updates & $\approx55$ min & $\approx35$ min \\
\bottomrule
\end{tabularx}
\end{table}

The benefits depend primarily on whether generation can supply data continuously. Under a short-problem workload, the complete-response approach has a median wait of approximately 0.3 seconds, and its queue is usually nonempty. When long-generation tasks are added, this value rises to approximately 78--87 seconds, and the queue is more prone to depletion. Segmented delivery is therefore better suited to workloads with longer generations or greater variation in latency across samples. Its value lies in mitigating uneven supply, not in replacing all complete-response pipelines.

A separate historical timing summary has an unresolved conversion unit and is listed in Appendix~\ref{app:statistics}. It is excluded from calculations of throughput ratios or end-to-end speedup for synchronous continuation.

\subsection{Overall Cost: Target Quality and the Choice of Switching Point}\label{sec:cost}

The synchronous continuation quality trajectory and the separate prototype timings support, respectively, the capacity for recovery along the continuation path and the systems value of finer-grained delivery. Benefits for the entire path, however, must be assessed by the total training time required to reach a target quality.

\paragraph{Continuation cost depends on the quality target.}
If the target quality is already attained during the asynchronous phase, no subsequent synchronous cost is necessary. Otherwise, total runtime must include switching overhead and continuation, including the plateau during which quality has not yet clearly improved. Faster updates therefore do not guarantee faster attainment of target quality. We currently lack complete time curves under common resources, consistent timing units, and paired evaluations, and thus cannot quantify the net speedup for different quality targets.

\paragraph{The switching point is a control variable to be optimized.}
An ``asynchronous first, synchronous later'' strategy uses existing weights and the consumption interface to improve quality without changing the student, teacher, or distillation objective. This study switches only at total updates $u=352$ and does not search for the optimal switching point. Whether an earlier switch reduces recovery cost, and whether gradual adjustments to freshness or position coverage can replace a direct switch, remain open questions.

Together, the two levels of experiments support treating the consumption protocol as an integrated stage-wise design: repairing the supervision quantity establishes basic training conditions, subsequent protocol adjustments continue to exploit existing weights, and the timing of those adjustments is determined jointly by the quality trajectory and total cost, rather than by assuming that a particular local rule is always optimal.

\section{Discussion and Conclusions}\label{sec:limitations}

\subsection{Decoupling Changes Training Organization, Not the Local Objective}

The distinguishing mechanism is that successive new segments from the same response continue to enter the training candidate pool, rather than supplying only a single truncated prefix or using continuation solely for probing. Local objectives, masking, and asynchronous execution each have precedents. Organizing them into an interface for repeated within-response delivery and independent trainer-side batching is what enables completion, delivery, and updates to be designed separately. The associated cost is also clear: old prefixes preserve the correct conditioning context but repeatedly incur computation. Whether earlier delivery is worthwhile must be assessed jointly in terms of quality and time under the specific workload.

Consumption diagnostics further show that delivery flexibility requires corresponding organization of the gradients. Quantity, position coverage, and version jointly characterize the supervision entering updates; a single disagreement score or tail fraction cannot replace the full consumption distribution. Positive results from the supervision budget, local peak gains from freshness filtering and disagreement--overlap selection, and the remaining limitations of selection and local-objective interventions support diagnosing quality costs through the consumption process rather than attributing gaps of different origins to one mechanism. Synchronous continuation jointly adjusts subsequent consumption and optimizer state. Matched optimizer-reset controls and per-position consumption records would help further distinguish the mechanisms of recovery.

Policy-gradient research offers related insights into feedback granularity. For sequence-level rewards, GSPO~\cite{gspo} uses a length-normalized sequence importance ratio to mitigate the accumulation of high-variance noise from tokenwise weighting in long sequences. Our focus is instead on how per-position distillation supervision enters updates under prefix dependence and version lag. These are not the same failure mechanism, but both suggest that training organization should account for feedback granularity and contextual dependence, rather than judging update effects from the local loss alone. Regeneration with current weights can refresh sampled prefixes and is not unique to distillation. In our short-horizon screening, it has the highest method mean among the tested interventions, 62.70\%, still 4.09 percentage points below the synchronous anchor; unequal run counts do not establish its scheduling advantage.

\subsection{Consumption Precision as a Stage-Wise Control Variable}

Curriculum and self-paced learning incorporate training progress into the organization of examples~\cite{curriculum,selfpaced}; we examine how the consumption rules for the same type of supervision stream can change across stages, without requiring a different student, teacher, or newly devised distillation objective. This curriculum perspective centered on consumption precision adjusts how supervision enters updates, without presupposing harder examples or stricter filtering later in training. Because the student generates data online, retaining the same data source does not imply that both stages visit exactly the same prefixes. Nor do synchronous and asynchronous operation constitute two fixed levels of consumption precision.

Aligning supervision quantity yields clear benefits, while freshness filtering and disagreement--overlap selection improve local peaks without establishing sustained advantages across the candidate window. This indicates that the value of consumption control lies not in continually increasing filtering strength, but in coordinating supervision quantity, coverage, and update organization. The budget remedy and synchronous continuation support the feasibility of stage-wise adjustment, but do not justify partitioning learning into mutually exclusive phases in which ``only quantity matters early and only consumption precision matters later.'' Most freshness and position-selection experiments begin late in training, so they cannot establish that the composition of early supervision is irrelevant. Failure of the tested interventions to consistently close the residual gap also does not imply that the asynchronous pool lacks all effective supervision or that switching to synchronous consumption is the only route to recovery.

The goal of stage-wise design is therefore not to prescribe a fixed ``coarse-to-fine'' curriculum in advance, but to adjust coverage, version control, and consumption scheduling according to quality targets while meeting the actual supervision budget. The asynchronous-first, synchronous-later approach studied here is one feasible path. Which controls are needed at which stages, and whether gradual transitions can reduce recovery costs, must be determined jointly from consumption statistics and quality trajectories. Decoupling the three clocks provides the interface for these choices without predetermining the optimal choice.

\subsection{Scope and Basis of Comparison}

The quality experiments use a synchronous baseline as their reference and cover one model pair, one engine, and one seed lineage. Comparisons with closely related methods such as TOPD and ADWIN focus on mechanisms; we do not reproduce their performance under a unified configuration. Historical diagnostics are reported separately according to their respective configurations and observation windows, and their MATH500 evaluation coverage and answer-extraction procedures are distinguished from those of the main experiments. Cross-model and multiseed validation, together with joint quality--time comparisons, are the next steps in testing generality. This study examines quality recovery and data-supply benefits separately and does not combine them into a claim of quality-preserving acceleration for the stage-wise strategy.

Primary quality scores are test-set maxima within common candidate checkpoints and are subject to selection optimism, not independent test estimates after validation-based selection. Matching candidate ranges, evaluation opportunities, and update ceilings does not equate the selected updates or GPU costs. Each main-window method has one trajectory; window means and full curves describe its correlated evaluation process, not seed variation. Consumption audits and prototype performance retain the precision of the summary records; detailed conventions appear in Appendix~\ref{app:records}. Historical records do not fully specify training-data counts by source, postfiltering dataset sizes, or the final configuration of every run. We therefore distinguish launch recipes from actual run conditions. Records without update mappings are reported descriptively and excluded from common-candidate comparisons.

\subsection{Conclusions}

This paper grounds streaming OPD in the decoupling of three clocks: local supervision need not wait for response completion, segment delivery need not directly trigger a parameter update, and consumption rules need not remain fixed throughout training. Complete causal prefixes and new-segment masks make each response a continuing source of training data, while the consumption protocol organizes that supervision into updates that can be adjusted independently and across stages.

The experiments show that this flexibility requires explicit consumption control.
Supervised-budget matching reduces the early best-score gap from 4.55 to 2.42
percentage points within seven common candidates. Synchronous continuation
jointly resets the optimizer and switches consumption from the same
asynchronous checkpoint. Over the seven common candidates
$u=384,416,\ldots,576$, it reaches 73.39\%, compared with 74.30\% for the
synchronous baseline and 71.95\% for supervised-budget matching, narrowing
the best-score gap from 2.35 to 0.91 percentage points. The unpaired
eleven-point extended record reaches 76.12\%, without a contemporaneous
synchronous reference. These observed test-set maxima are subject to selection
optimism; means and full curves remain process supplements. Freshness
filtering and disagreement--overlap selection improve local peaks but do not
establish sustained advantages or consistently close the residual gap. Low
scores from tail-quota selection and vocabulary-support interventions remain
unpaired supplementary observations. Reduced data-supply waiting in a separate
prototype demonstrates the systems value of finer-grained delivery. Because
the completion counts in the historical timings are not yet aligned, the net
benefit of the stage-wise strategy still requires accounting for the full cost
of reaching target quality.

This shifts the research question from whether training should remain asynchronous throughout to how available supervision should be consumed at different stages. Segmented delivery expands the available choices, the consumption protocol organizes gradients, and stage-wise adjustment connects quality and efficiency targets. Asynchrony is one scheduling option within this design, not a training objective that must persist throughout the entire process.

\appendix
\section{Supplementary Evaluation Records and Statistical Definitions}\label{app:records}

\subsection{Training Counts and Cross-Machine Reevaluation}\label{app:counting}

\paragraph{Method names and archival identifiers.}
The main-text names map to the archived identifiers as follows: synchronous
baseline (\texttt{base\_s1}); naive streaming (\texttt{seal\_s1}); raw-token budget
(\texttt{tok80k}); supervised-budget matching (A2, \texttt{supv80k}); reduced
learning rate (B, \texttt{lr58}); synchronous continuation (A3), which jointly
switches consumption and resets the optimizer; high-disagreement selection
(C, C-late); disagreement--overlap selection (T, T-late); uniform random
selection (C0, \texttt{rand}); freshness filtering (\textsc{FreshOnly});
teacher-support mass reweighting (D-late, \texttt{mass}); tail-quota selection
(\texttt{comp}); dynamic vocabulary truncation (\texttt{dynk}); and
bidirectional rank divergence (\texttt{dbild}). The scheduling-screen mappings
are synchronous baseline (\texttt{base\_anchor}, s2), naive streaming
(\texttt{seal\_anchor}, s2), completeness gating (\texttt{seal\_cc}, s1),
terminal-first scheduling (\texttt{seal\_tf}, s1), stale-segment replay
(\texttt{seal\_rp}, s1/s2), and regeneration with current weights
(\texttt{seal\_rg}, s1/s2). Supervised-budget matching retains the same name
when training continues; branch codes below serve provenance only and do not
define additional methods. Historical naive-sealing and naive-segmentation
names refer to naive streaming within their respective configuration groups.

\paragraph{Directories and the update axis.}
In a checkpoint directory named \texttt{global\_step\_}$g$, $g$ is the broadcast parameter-version index. The main run broadcasts once every four trainer updates, with a regular checkpoint interval of eight parameter versions. The long-horizon control additionally includes a nonperiodic final checkpoint at $g=149$ ($u=596$). The log field \texttt{training/global\_step} directly records trainer updates and should not be multiplied by four. Synchronous continuation (A3) and the late-start position-selection experiments begin from the supervised-budget matching (A2) weights at $u=352$ and reset the optimizer. Sharing initial weights does not imply identical subsequent optimizer states or data trajectories. The identifier s1b denotes a separately reported continuation branch and is not merged with s1 into a continuous trajectory. Main comparisons use the fixed windows in Section~\ref{sec:setup}; supplementary branch records do not create additional method rows.

\paragraph{Reduced learning rate (B, \texttt{lr58}): supplementary checkpoints.}
The intervention retains naive streaming's sample-count batching and uses
$0.58$ times the default learning rate without a token budget. At $u=448$,
its accuracy is 69.22\%, compared with 71.49\% for supervised-budget matching.
At $u=596$, the corresponding scores are 72.02\% and 71.65\%. These records
preserve the low intermediate score and later recovery; they are not separate
main results or candidates for the seven-point early best score. The two
interventions have different intermediate trajectories and similar endpoint
performance, which does not establish a common causal mechanism.

\paragraph{Cross-machine reevaluation.}
For uniform random selection (\texttt{rand}), teacher-support mass reweighting
(\texttt{mass}), and freshness filtering (\textsc{FreshOnly}), 16 paired
reevaluations of the same checkpoints on a second evaluation machine have a
mean paired difference of $+0.01$, a mean absolute difference of $0.47$, and a
maximum absolute difference of $0.84$ percentage points. These records describe
cross-machine reevaluation differences and do not constitute a general
significance threshold. Similar scores or overlapping intervals do not
establish statistical equivalence.

\subsection{Detailed Evaluations of Position-Selection and Objective Interventions}\label{app:selection}

\begin{figure}[!htbp]
\centering
\includegraphics[width=\textwidth]{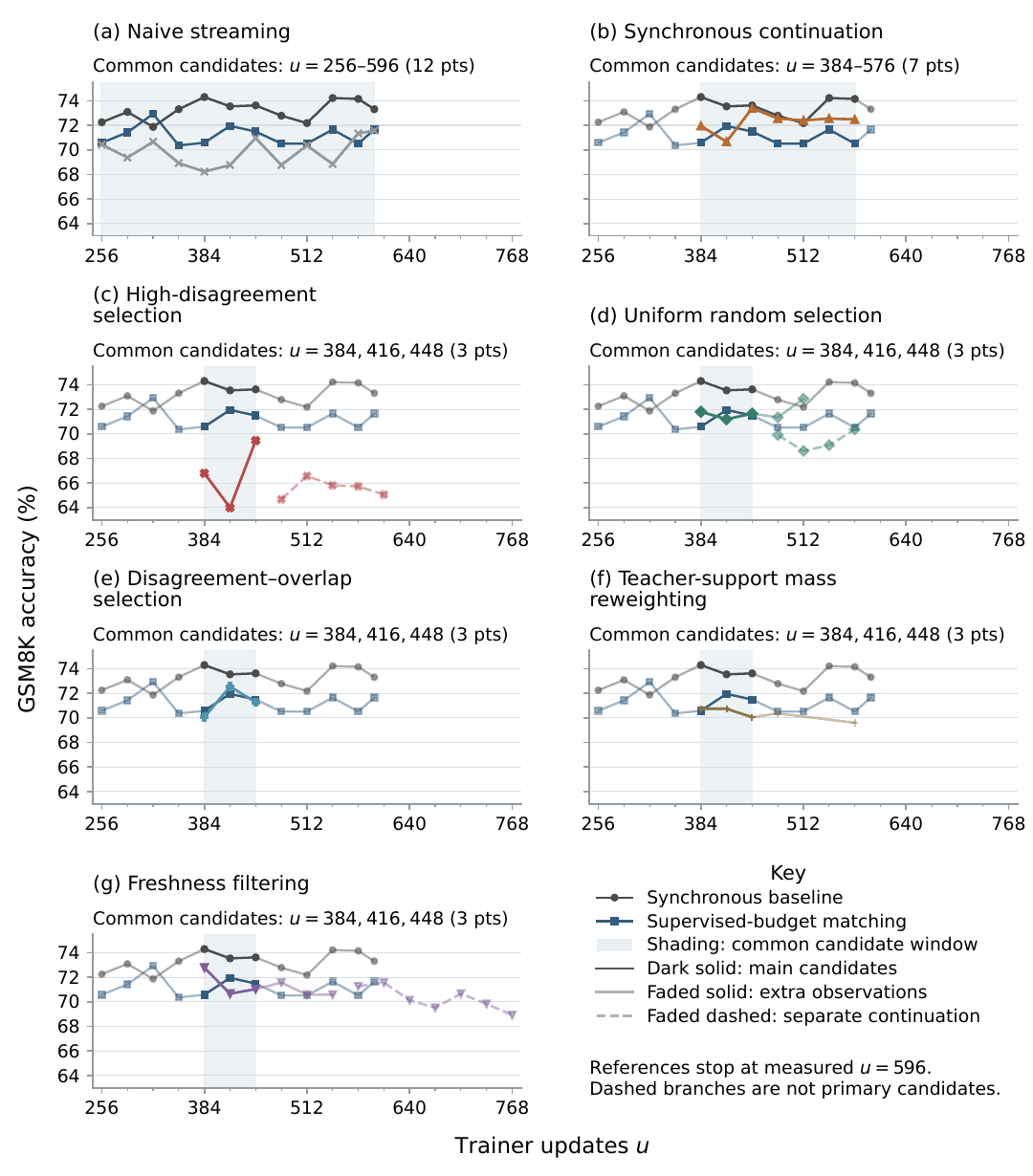}
\caption{Full trajectories in method-specific panels with shared axis scales.
Shading marks the primary candidate windows: twelve points from $u=256$ to
$596$ for naive streaming, seven from $u=384$ to $576$ for synchronous
continuation, and $u=384,416,448$ for the other interventions. Each panel
includes the synchronous baseline and supervised-budget matching as references.
Faded curves show supplementary records; dashed curves denote separate
continuation branches. Only measured points with known update mappings are
shown, without adding points or joining branches. Records outside the candidate
windows are excluded from primary checkpoint selection.}
\label{fig:curves-full}
\end{figure}

Table~\ref{tab:cells} selects each method's best test accuracy only from the
common candidates $u=384,416,448$. Here, all available supplementary values
and original ranges are retained,
including the separately identified initial (s1) and continuation (s1b) branches.
All values below are GSM8K accuracies (\%), listed in recorded order. Different
numbers of observations and different windows do not constitute uniform
replication. Figure~\ref{fig:curves-full} preserves the full trajectories with
known update mappings; it is supplementary, not a replacement for the
common-window comparisons in Figure~\ref{fig:curves}.

\paragraph{High-disagreement selection (C-late).}
There are eight checkpoints in total: the three s1 points at $u=384,416,448$
are 66.79, 63.99, and 69.45, and the five s1b points are 64.67, 66.57, 65.81,
65.73, and 65.05. The combined reported range is 63.99--69.45; both branches
remain below 70.00\%. The s1 best of 69.45\% at $u=448$ supplies the
common-candidate comparison; its 66.74\% mean is a process supplement.
The other branch remains supplementary. MATH500 endpoint
evaluations and length statistics are reported separately in Section~\ref{sec:select}.

\paragraph{Disagreement--overlap selection (T-late).}
The five points are 70.05, 72.55, 71.27, 71.27, and 71.27. The final three
are identical, and the range is 70.05--72.55. The first three common
candidates yield the best of 72.55\% at $u=416$ in Table~\ref{tab:cells},
0.60 percentage points above supervised-budget matching's best. Their
process mean is 71.29\%. The local peak improvement does not establish a
sustained advantage.

\paragraph{Uniform random selection (C0).}
The five s1 points are 71.80, 71.19, 71.65, 71.34, and 72.86, corresponding
to $u=384,416,\ldots,512$, with a range of 71.19--72.86. Each of the first
three exceeds its high-disagreement-selection counterpart; their best is
71.80\% at $u=384$ and their process mean is 71.55\%. The 72.86\% score at
$u=512$ is outside the three-candidate set and cannot enter its best-score
comparison. The s1b scores at $u=480,512,544,576$ are 69.90, 68.61, 69.07,
and 70.36, with a range of 68.61--70.36. The branches overlap at $u=480,512$
and are retained separately, not joined. The combined reported range is
68.61--72.86. The lower s1b scores remain separate records, not a second
method in the main comparison.

\paragraph{Freshness filtering (FreshOnly).}
The six s1 points cover $u=384,416,\ldots,544$ and are 72.78, 70.66, 71.04,
71.57, 70.58, and 70.58, with a mean of 71.20, compared with 71.12 for
supervised-budget matching over the same historical six-point window. The
nine s1b points are 71.27, 71.57, 70.13, 69.52, 70.66, 69.83, 68.92, 69.98,
and 69.75. The combined fifteen-point range is 68.92--72.78, with a pooled
mean of 70.59. This historical pooled mean describes only these records;
it is neither a seed mean nor a matched comparison against the six-point
supervised-budget mean. The later records do not sustain the earlier
improvement, but the lower uniform-random-selection s1b scores also preclude
attributing that change exclusively to freshness filtering. The main result
uses the best of 72.78\% at $u=384$ within the three common candidates,
0.83 percentage points above supervised-budget matching's best. Freshness
filtering wins only the first of these three points; the 71.49\% mean is
retained as a process supplement, not the primary score in Table~\ref{tab:cells}.

\paragraph{Teacher-support mass reweighting (D-late).}
Seven checkpoints cover $u=384$--$576$, with a range of 69.60--72.10.
Figure~\ref{fig:curves-full} shows five mapped points at $u=384$, 416, 448,
480, and 576, with values of 70.74, 70.74, 70.05, 70.36, and 69.60,
respectively. The first three yield a best of 70.74\% at both $u=384$ and
$416$; the earliest tie, $u^*=384$, is selected. Their process mean is 70.51\%.
The full-range high of 72.10\% lacks a pointwise update mapping and cannot
enter the main candidate set. The remaining range information is retained
without manufacturing missing pointwise mappings.

\paragraph{Tail-quota selection (\texttt{comp}).}
The five points are 64.06, 66.72, 68.01, 65.50, and 66.03, with a range of
64.06--68.01, a best of 68.01\%, and a process mean of 66.06\%. These low
scores are retained as an unpaired supplementary observation. Without a
reliable step map, the selected update is unknown and they do not enter
Table~\ref{tab:cells}, a common-candidate baseline difference, or a comparison
against uniform random selection's different coverage window.

\paragraph{Dynamic vocabulary truncation (\texttt{dynk}).}
The four points are 64.97, 68.39, 68.31, and 68.61, with a range of
64.97--68.61, a best of 68.61\%, and a process mean of 67.57\%. The final
three lie within 68.31--68.61. These records lack a reliable step map;
the selected update is unknown and no synchronous-baseline difference is
computed for this unpaired supplement.

\paragraph{Bidirectional rank divergence (\texttt{dbild}).}
The three points are 64.22, 63.31, and 62.93, after which the branch was
terminated. Their range is 62.93--64.22, their best is 64.22\%, and their
process mean is 63.49\%. Like dynamic vocabulary truncation, the selected
update is unknown and these records cannot supply a common-candidate
comparison or a synchronous-baseline difference.

\subsection{Short-Horizon Screening and Review of Historical Scoring}\label{app:diagnostics}

\paragraph{Run mapping for scheduling screening.}
Each record below is the accuracy at $u=16$ within the independent screen.
The identifiers s1 and s2 are run IDs only and do not establish independent
random seeds. Synchronous baseline, naive streaming, completeness gating,
and terminal-first scheduling each have one run; stale-segment replay and
regeneration with current weights each have two, with s2 added subsequently.
Every difference uses the same synchronous anchor of 66.79\%, not an
individually paired run. Within-run selection has only the candidate $u=16$;
Table~\ref{tab:timing} then averages those selected values with equal run
weights, rather than picking the best run or treating additional records as
new methods. Differences are computed after averaging, with final values
rounded to two decimal places. Run counts are unequal and do not constitute
a matched-seed design; these screening results do not establish a stable
ranking or statistical significance.
Entries below report accuracy (\%), followed by the baseline difference
in parentheses (percentage points).
\begin{itemize}
  \setlength{\itemsep}{2pt}
  \setlength{\parsep}{0pt}
  \item Synchronous baseline (\texttt{base\_anchor}, s2): 66.79 ($0.00$).
  \item Naive streaming (\texttt{seal\_anchor}, s2): 59.06 ($-7.73$).
  \item Completeness gating (\texttt{seal\_cc}, s1): 61.79 ($-5.00$).
  \item Terminal-first scheduling (\texttt{seal\_tf}, s1): 53.15 ($-13.64$).
  \item Stale-segment replay (\texttt{seal\_rp}): s1, 58.68 ($-8.11$);
        s2, 52.84 ($-13.95$); mean, 55.76 ($-11.03$).
  \item Regeneration with current weights (\texttt{seal\_rg}):
        s1, 64.82 ($-1.97$); s2, 60.58 ($-6.21$); mean, 62.70 ($-4.09$).
\end{itemize}

\paragraph{High-lag filtering.}
The high-lag summary after scoring was standardized contains two sets of records. One lists 67.40\% for retaining all supervision and 61.03\% for freshness filtering; a subsequent point for the freshness-filtered branch is 66.87\%, without a corresponding unfiltered score. The other set has endpoint values of 40.56\% for retaining all supervision and 60.80\% for freshness filtering, reversing the ordering. The summary does not fully specify run identifiers and checkpoint mappings. We therefore neither calculate paired effects nor infer which lag configuration is better suited to filtering.

\paragraph{Answer-extraction correction.}
After correction of answer-marker extraction, the accuracies of six early
runs are 68.70, 67.40, and 61.50\% for sealed-segment delivery and 70.70,
68.80, and 69.10\% for complete-response delivery. These two-decimal displays
retain the historical source's one-decimal precision. The earlier judgment
that ``four of six runs collapsed'' arose from a scoring defect. One
complete-response record ends at an earlier checkpoint, so these data serve
to correct the collapse judgment rather than estimate stability at matched
endpoints. Degenerate outputs in the PG diagnostics persist after rescoring
and should be distinguished from scoring errors and ordinary checkpoint
fluctuations.

\subsection{Synchronous Continuation (A3): Records and Synchronous References}\label{app:continuation}

\paragraph{Synchronous continuation (A3): supplementary late-window analysis.}
The complete seven-point main candidate set ($u=384,416,\ldots,576$) gives
best test accuracies of 74.30\% at $u=384$, 71.95\% at $u=416$, and 73.39\%
at $u=448$ for the synchronous baseline, supervised-budget matching, and
synchronous continuation, respectively. The latter two best-score differences
are $-2.35$ and $-0.91$ percentage points. As process statistics over these
seven points, their means are 73.54\%, 71.03\%, and 72.28\%; the latter two
mean differences are $-2.51$ and $-1.26$ percentage points, corresponding
to approximately 50\% recovery of the \emph{mean} gap.
For the supplementary late five points ($u=448,480,\ldots,576$), the same
methods average 73.39\%, 70.93\%, and 72.67\%; the latter two mean differences
are $-2.46$ and $-0.72$ percentage points, or approximately 71\% recovery
of the \emph{mean} gap. Both percentages are process summaries, not recovery
percentages for the best-score gap. The five-point subwindow excludes the
first two post-switch evaluations and describes the later trajectory, not
an established steady state or an alternative main comparison. Continuation
still trails the baseline by 1.67 percentage points at each of the final two
checkpoints. Both windows describe the joint schedule change and optimizer
reset, not the isolated effect of either.

\paragraph{Synchronous continuation (A3): restart position and counts.}
After its initial segment was interrupted, synchronous continuation was
restarted from the checkpoint indexed $g=56$. Cumulative stage updates are
$r=4g=224$, and total updates are $u=352+r=576$; the restart does not reset
the stage count to zero. The first entry in the extended record is 72.40\%,
whereas the earlier segment records 72.48\% at the same count. Both records
are retained separately. This discrepancy is not interpreted as an effect of
restarting, nor are the records spliced into a seamless paired trajectory.

\paragraph{Synchronous continuation (A3): extended pointwise evaluations.}
The eleven extended candidates at $r=224,256,\ldots,544$ correspond to
$u=576,608,\ldots,896$. Their best observed test accuracy is 76.12\% at
$r=480$ ($u=832$), and their overall arithmetic mean is 74.06\% as a process
supplement. The four points at $r=224$, 256, 288, and 320 are 72.40, 72.48, 72.48,
and 72.48\%, with a mean of 72.46\% and a range of 0.08 percentage points.
The two points at $r=352$ and 384 are 75.66\% and 74.98\%, and the five at
$r=416$, 448, 480, 512, and 544 are 74.30, 74.83, 76.12, 74.60, and
74.30\%. The latter seven points have a mean of 74.97\% and a range of
74.30--76.12\%. The first-four and later-seven summaries describe the
plateau-to-improvement pattern; neither replaces the full eleven-point mean.
These correlated observations are not independent experimental replications
and have no contemporaneous synchronous reference. They are not pooled
with the main seven-point matched comparison.

\begin{figure}[!htbp]
\centering
\includegraphics[width=\textwidth]{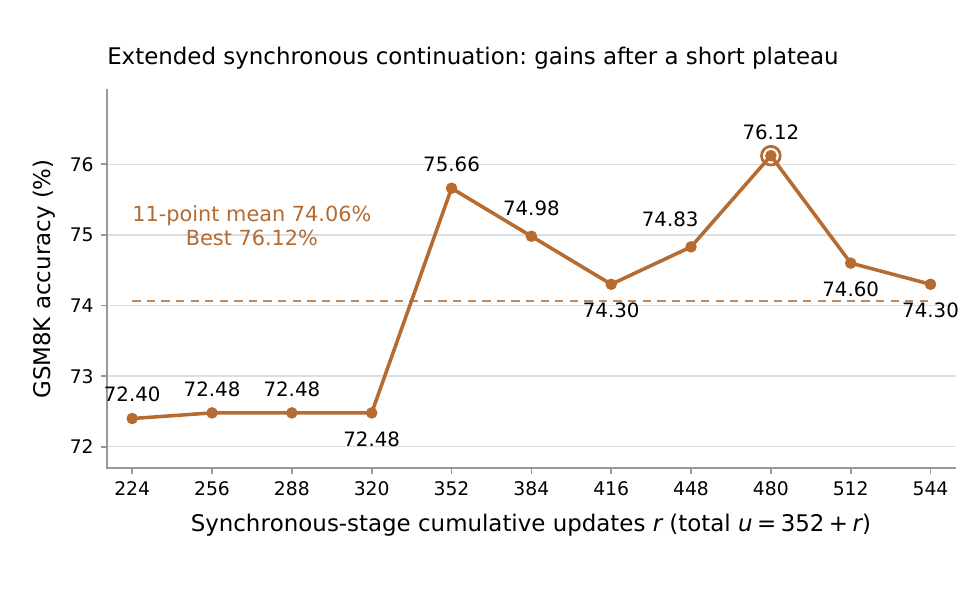}
\caption{Supplementary extended synchronous continuation. Cumulative stage
updates $r$ and total updates satisfy $u=352+r$, with the restart at $r=224$.
The open circle highlights the best test accuracy of 76.12\% at $r=480$
($u=832$) within all eleven candidates $r=224,256,\ldots,544$. The dashed
line indicates their process mean of 74.06\%, including the first-four-point
plateau; it is not a confidence interval. Connecting lines are not smoothed.
This unpaired figure is not spliced with Figure~\ref{fig:a3-paired} and has
no contemporaneous synchronous reference.}
\label{fig:a3-extended}
\end{figure}

\paragraph{Synchronous continuation (A3): apparent latency.}
The first four observations span 96 updates. A clearly higher accuracy is
first observed after another 32 updates, placing that observation 128 updates
after the start of the extended record and 352 updates after the initial
switch to synchronous consumption. The first higher point at $r=352$
corresponds to total updates $u=704$. This apparent latency indicates that
benefits may appear later in the evaluation curve; a short window without
sustained improvement does not rule out further recovery. Sparse checkpoints
cannot identify the precise onset of improvement or a generally applicable
waiting period. The trajectory is not monotonically increasing, and the
plateau-to-improvement pattern does not establish its cause.

\paragraph{Synchronous baseline: records outside the paired window.}
Six additional synchronous-baseline points after $u=596$ are 73.24, 73.31,
69.52, 74.30, 75.28, and 73.62\%. They exhibit fluctuations rather than a
fixed plateau and are excluded from the synchronous-continuation matched
window. A historical long-horizon synchronous branch
(\texttt{best\_base\_long\_fkl}) also has a recorded evaluation of 77.41\%,
above the extended synchronous-continuation peak of 76.12\%. These values
come from different runs and coverage windows. They cannot substitute for
an equal-update comparison or establish the capability ceilings of the two
approaches.

\subsection{Summary Statistics and Timing Definitions}\label{app:statistics}

\paragraph{Aggregation of results.}
The fixed candidate sets and selection rule are given in Section~\ref{sec:setup}.
Each method independently selects its maximum test accuracy within the same
candidate set, breaking ties at the earliest $u^*$. Best-score differences
compare those maxima, not same-update paired scores. Where multiple runs
are available, selection occurs within each run before equal-weight averaging
of the selected scores. The independent timing screen has only the candidate
$u=16$ per run, so its averages are unchanged; unequal run counts are not
matched seeds and no best run is selected. The other main-window comparisons
have one trajectory per method in each group.

Tables~\ref{tab:checkpoint-audit-early-long} and~\ref{tab:checkpoint-audit-late-cont}
audit the selected updates and retain arithmetic window means as process
supplements. Mean $\Delta$ is relative to the synchronous mean in the same
group, computed from unrounded values before final two-decimal rounding.
The means summarize correlated checkpoints, not seeds; neither they nor
test-set maxima establish statistical equivalence. Maximizing test accuracy
has selection optimism and is not validation-based selection followed by
independent testing. Candidate ranges, evaluation opportunities, and update
ceilings are aligned, not necessarily the selected updates or actual GPU costs.
Early reduced-learning-rate results (B) are based on pointwise summary values.
Unmapped and unpaired supplements do not enter common-candidate baseline
differences, and historical subwindows do not replace the primary sets.

\begin{table}[!htbp]
\centering\small
\setlength{\tabcolsep}{4pt}
\caption{Checkpoint-selection audit for the early and long-horizon comparisons.
Early: seven candidates $u=32,64,\ldots,224$, excluding $u=256$.
Long horizon: twelve candidates $u=256,288,\ldots,576$ plus $596$.
Each method has one trajectory per group; means describe checkpoints, not
seeds. Mean $\Delta$ uses the synchronous mean within the same group.}
\label{tab:checkpoint-audit-early-long}
\begin{tabularx}{\textwidth}{Yrrrr}
\toprule
Method & $u^*$ & Best (\%) & Mean (\%) & Mean $\Delta$ (pp) \\
\midrule
\multicolumn{5}{l}{\textit{Early: seven candidates}} \\
Synchronous baseline & 192 & 73.84 & 71.55 & $0.00$ \\
Naive streaming & 192 & 69.29 & 67.74 & $-3.80$ \\
Raw-token budget & 160 & 70.96 & 68.86 & $-2.69$ \\
Supervised-budget matching & 128 & 71.42 & 69.65 & $-1.90$ \\
Reduced learning rate & 160 & 71.49 & 69.90 & $-1.65$ \\
\midrule
\multicolumn{5}{l}{\textit{Long horizon: twelve candidates}} \\
Synchronous baseline & 384 & 74.30 & 73.22 & $0.00$ \\
Naive streaming & 596 & 71.57 & 69.85 & $-3.37$ \\
Supervised-budget matching & 320 & 72.93 & 71.18 & $-2.04$ \\
\bottomrule
\end{tabularx}
\end{table}
\FloatBarrier

\begin{table}[!htbp]
\centering\small
\setlength{\tabcolsep}{4pt}
\caption{Checkpoint-selection audit for late-start interventions and synchronous
continuation. Late start: three candidates $u=384,416,448$.
Continuation: seven candidates $u=384,416,\ldots,576$.
Each method has one trajectory per group; means describe checkpoints, not
seeds. Mean $\Delta$ uses the synchronous mean within the same group.}
\label{tab:checkpoint-audit-late-cont}
\begin{tabularx}{\textwidth}{Yrrrr}
\toprule
Method & $u^*$ & Best (\%) & Mean (\%) & Mean $\Delta$ (pp) \\
\midrule
\multicolumn{5}{l}{\textit{Late start: three candidates}} \\
Synchronous baseline & 384 & 74.30 & 73.82 & $0.00$ \\
Supervised-budget matching & 416 & 71.95 & 71.34 & $-2.48$ \\
High-disagreement selection & 448 & 69.45 & 66.74 & $-7.08$ \\
Disagreement--overlap\newline selection & 416 & 72.55 & 71.29 & $-2.53$ \\
Uniform random selection & 384 & 71.80 & 71.55 & $-2.27$ \\
Freshness filtering & 384 & 72.78 & 71.49 & $-2.33$ \\
Teacher-support mass\newline reweighting & 384 & 70.74 & 70.51 & $-3.31$ \\
\midrule
\multicolumn{5}{l}{\textit{Synchronous continuation: seven candidates}} \\
Synchronous baseline & 384 & 74.30 & 73.54 & $0.00$ \\
Supervised-budget matching & 416 & 71.95 & 71.03 & $-2.51$ \\
Synchronous continuation\newline (optimizer reset) & 448 & 73.39 & 72.28 & $-1.26$ \\
\bottomrule
\end{tabularx}
\end{table}
\FloatBarrier
Teacher-support mass reweighting reaches the same 70.74\% peak at $u=384$
and $416$; the table records the earliest tie, $u^*=384$.

Consumption audits and prototype performance retain the precision and
statistical definitions of their source records rather than being converted
to accuracy means. These summaries do not substitute for verification of
per-question evaluations or per-step timings. Training-data counts by source,
postfiltering sizes, and the final effective configurations of individual
runs are not fully recorded, so launch recipes and actual run conditions are
described separately.

\paragraph{Record granularity.}
Short-horizon scheduling screening and the two sets of PG diagnostics can be
checked against per-question scoring files containing 1319 questions per set.
Best scores and process means for the early, long-horizon, late-start, and
common-candidate synchronous-continuation comparisons are calculated from
per-checkpoint summaries. Extended synchronous continuation is based on the
listed pointwise evaluation records, from which its maximum and means can
be recalculated; this does not imply that per-question outputs have been
independently checked. Complete per-question or per-pair originals are
unavailable for the reduced learning rate's $u=596$ endpoint and the 16-pair
cross-machine reevaluation statistics. These are retained as historical
summaries, not as evidence of independent reproduction. Replotting or
aggregation does not give records at different levels the same evidentiary
strength.

\paragraph{Consumption-side audits.}
The source summary for Table~\ref{tab:audit} does not specify the start and end steps or interval-aggregation method for each statistic. Ranges such as 46--48K are therefore not interpreted as extrema across all update steps or as confidence intervals. The fraction of supervision from terminal segments is calculated using segment flags and is not the fraction of individual EOS tokens. The first two rows of Table~\ref{tab:posaudit} summarize per-step quantiles over the full candidate pool for the first 8 steps, rather than quantiles recalculated after pooling all positions. The final two rows summarize fresh and stale position groups at approximately step 8. KL and mass characterize local distributions and do not directly measure training benefit.

\paragraph{Separate prototype statistics.}
The rows of Table~\ref{tab:system} do not come from a single measurement window. The symbol $\approx$ retains approximate precision, and intervals are the ranges reported in the source records. The symbol $\leq$ denotes an upper bound given in the summary, not a theoretical guarantee or a maximum verified against per-step data. Complete-response delivery can also operate asynchronously; less supply waiting does not imply proportionally shorter runtime.

The historical prototype targeted 64 updates. Although the prose summary states that both runs completed, the completion field lists 64/64 for sealed-segment delivery and 63/64 for complete-response delivery. Subsequent records targeting 128 updates likewise list 128/128 and 127/128, respectively. Without the corresponding per-step timings, it is impossible to determine whether the discrepancy arises from indexing conventions or the actual number completed. We therefore retain the runtimes of approximately 35 and 55 minutes, whose ratio is approximately 1.6, without treating it as a verified equal-update or equal-quality speedup.

The sealed-segment run targeting 64 updates dropped 499 segments because of queue overflow. The source record reports approximately 2.9\% using a denominator of approximately 17K, while the complete-response control reports zero drops. The subsequent sealed-segment run targeting 128 updates increased queue capacity fourfold and reported zero drops. This configuration change cannot retroactively replace the segment-drop conditions of the earlier timing comparison.

\paragraph{A separate summary of update cadence.}
A historical timing summary using the same teacher--student pair and 64-sample batching (mini-batch $16\times4$ batches) derives 4.0--4.4 minutes for asynchronous operation and approximately 18 minutes for synchronous operation from checkpoint timestamps. The summary does not establish whether the conversion unit is a trainer update or a parameter-version cycle. Identical models and batching also do not imply identical GPU configurations. We therefore retain these time ranges without reporting a per-update throughput ratio, and do not combine them with the separate prototype timings to estimate end-to-end acceleration for synchronous continuation.

\end{document}